\documentclass{article} 
\usepackage{colm}
\usepackage[colorlinks = true,
            linkcolor = blue,
            urlcolor  = blue,
            citecolor = blue,
            anchorcolor = blue]{hyperref}   
\usepackage[utf8]{inputenc} 
\usepackage[T1]{fontenc}    
\usepackage{url}            
\usepackage{booktabs}       
\usepackage{amsfonts}       
\usepackage{nicefrac}       
\usepackage{microtype}      
\usepackage{xcolor}         
\usepackage{fvextra} 
\usepackage{tabularx}
\usepackage{algorithm}
\usepackage{algorithmic}

\usepackage{microtype}
\usepackage{arydshln}
\usepackage{graphicx}
\usepackage{subcaption}
\usepackage{booktabs} 
\usepackage{tikz}
\usepackage{tcolorbox}
\tcbuselibrary{skins}
\usepackage{nicefrac}  
\usepackage{nicematrix}
\usepackage{multirow}
\usepackage[normalem]{ulem}
\useunder{\uline}{\ul}{}
\usepackage{wrapfig}

\usepackage{booktabs}
\usepackage{multirow}
\usepackage{tabularx}
\usepackage{makecell}
\usepackage{amssymb}
\usepackage{pifont}
\usepackage{amsmath}
\usepackage{tcolorbox}
\usepackage{listings}
\usepackage{xcolor}
\tcbuselibrary{skins}
\usepackage{hyperref}
\usepackage{float}

\usepackage{bbm}

\definecolor{disblue}{HTML}{2F80ED}
\definecolor{genorange}{HTML}{F2994A}
\definecolor{rerankpurple}{HTML}{7B61FF}
\definecolor{cotred}{HTML}{EB5757}
\definecolor{judgegreen}{HTML}{219653}

\newcommand{\tok}[2]{\textcolor{#1}{\texttt{#2}}}

\newcommand{\disemb}{\tok{disblue}{<dis\_emb>}}
\newcommand{\genemb}{\tok{genorange}{<gen\_emb>}}

\newcommand{\rerankthink}{\tok{rerankpurple}{<rerank\_think>}}
\newcommand{\reranklist}{\tok{rerankpurple}{<rerank\_list>}}
\newcommand{\rerankjudge}{\tok{judgegreen}{<rerank\_judge>}}

\newcommand{\cotfocus}{\tok{cotred}{<cot\_focus>}}
\newcommand{\cotanswer}{\tok{cotred}{<cot\_answer>}}

\usepackage{fontawesome}

\usepackage{bbding}
\makeatletter
  \newcommand\figcaption{\def\@captype{figure}\caption}
  \newcommand\tabcaption{\def\@captype{table}\caption}
\makeatother

\usepackage[utf8]{inputenc} 
\usepackage[T1]{fontenc}    
\usepackage{url}            
\usepackage{booktabs}       
\usepackage{amsfonts}       
\usepackage{nicefrac}       
\usepackage{microtype}      
\usepackage{array}
\newcolumntype{C}[1]{>{\centering\arraybackslash}m{#1}}
\usepackage{xcolor}         
\usepackage{color, colortbl}
\definecolor{citecolor}{HTML}{2980b9}
\definecolor{linkcolor}{HTML}{c0392b}
\definecolor{darkorange}{HTML}{FF8C00}
\definecolor{chocolate}{HTML}{D2691E}
\definecolor{darkgreen}{HTML}{006400}
\definecolor{darkblue}{HTML}{00008B}
\definecolor{mediumblue}{HTML}{0000CD}
\definecolor{dodgerblue}{HTML}{1E90FF}
\definecolor{royalblue}{HTML}{4169E1}
\definecolor{shadecolor}{RGB}{237,237,237}
\definecolor{backred}{RGB}{255, 190, 190}
\definecolor{backblue}{RGB}{210, 230, 250}

\definecolor{zrrgreen}{HTML}{008000}
\definecolor{zrrblue}{HTML}{4682B4}
\definecolor{zrrred}{HTML}{B22222}

\usepackage{amsmath}
\usepackage{graphicx}

\usepackage{multirow}
\usepackage{makecell}
\usepackage{caption}

\usepackage{adjustbox}
\usepackage{wrapfig}

\usepackage{float}
\usepackage{subcaption}

\usepackage{tikz}
\usepackage{tcolorbox}
\tcbuselibrary{skins}

\usepackage{longtable}

\usepackage[nosfdefault]{comicneue} 
\usepackage{booktabs}       
\usepackage{tabularx}
\usepackage{makecell}
\usepackage{array}

\newcolumntype{C}[1]{>{\centering\arraybackslash}m{#1}}

\usepackage{listings}
\usepackage{subcaption}
\usepackage{CJKutf8}
\usepackage{wasysym}
\newcommand{\huggingface}{\raisebox{-1.5pt}{\includegraphics[height=1.05em]{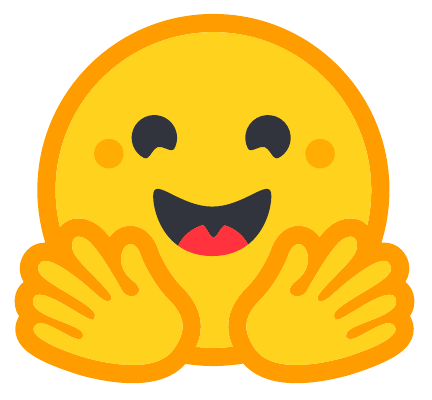}}\xspace}
\newcommand{\github}{\raisebox{-1.5pt}{\includegraphics[height=1.05em]{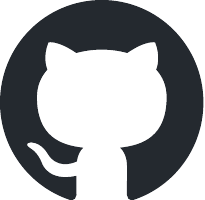}}\xspace}


\usepackage{framed}
\usepackage{makecell}
\usepackage{listings}
\definecolor{lightgray}{rgb}{.9,.9,.9}
\definecolor{darkgray}{rgb}{.4,.4,.4}
\definecolor{purple}{rgb}{0.65, 0.12, 0.82}
\lstdefinelanguage{JavaScript}{
  keywords={break, case, catch, continue, debugger, default, delete, do, else, false, finally, for, function, if, in, instanceof, new, null, return, switch, this, throw, true, try, typeof, var, void, while, with},
  morecomment=[l]{//},
  morecomment=[s]{/*}{*/},
  morestring=[b]',
  morestring=[b]",
  ndkeywords={class, export, boolean, throw, implements, import, this},
  keywordstyle=\color{blue}\bfseries,
  ndkeywordstyle=\color{darkgray}\bfseries,
  identifierstyle=\color{black},
  commentstyle=\color{purple}\ttfamily,
  stringstyle=\color{red}\ttfamily,
  sensitive=true
}
\usepackage{xspace}
\usepackage[shortlabels]{enumitem}

\newcommand\blfootnote[1]{%
  \begingroup
  \renewcommand\thefootnote{}\footnote{#1}%
  \addtocounter{footnote}{-1}%
  \endgroup
}
\usepackage{mdframed}
\newmdenv[
  topline=false,
  bottomline=false,
  rightline=false,
  leftline=true,
  linecolor=gray,
  linewidth=3pt,
  backgroundcolor=gray!10,
  skipabove=10pt,
  skipbelow=10pt,
  innertopmargin=5pt,
  innerbottommargin=5pt,
  innerleftmargin=10pt,
  innerrightmargin=10pt
]{promptblock}

\usepackage{wrapfig}
\usepackage{float}
\usepackage{xcolor}
\definecolor{kc}{rgb}{0.09, 0.45, 0.27}

\definecolor{lighttan}{rgb}{0.97,0.90,0.85}

\definecolor{CreamyBrown1}{HTML}{DBCCBD}
\definecolor{CreamyBrown2}{HTML}{F6EEE1}

\title{Learning from Failures: Retrieval-Centric CoT via Hard Negatives for Unified Multimodal Retrieval}

\author{Zelong Sun$^{*}$, Jun Wang$^{*}$, Kaicheng Yang$^{\dagger}$, Tiancheng Gu, Ziyong Feng, \textbf{Zhiwu Lu$^{\ddagger}$}\\
\\
 \github \textbf{GitHub:}  {\url{https://github.com/deepglint/UniME-R1}} \\
\huggingface  \textbf{HuggingFace:} \url{https://huggingface.co/DeepGlint-AI/UniME-R1}
}

\colmfinalcopy 

\begin{document}

\maketitle

\blfootnote{$^{*}$ Equal Contribution. $^{\ddagger}$ Corresponding Author. $^\dagger$ Project Leader.} 

\vspace{-0.5cm}

\begin{abstract}
Unified multimodal retrieval aims to identify candidates that satisfy complex user intent expressed through heterogeneous inputs. 
Although Large Vision-Language Model (LVLM)-based retrievers are efficient and scalable, directly encoding raw multimodal inputs often misses fine-grained discriminative cues, leading to confusion among semantically similar candidates.
Recent methods mitigate this limitation by generating Chain-of-Thought (CoT) rationales to enrich the query representation. However, such reasoning is typically derived from the query alone: \emph{it explains what the query describes, but not what the retriever misunderstands}. We argue that effective retrieval reasoning should instead be conditioned on retrieval feedback. 
Based on this insight, we introduce \textbf{UniME-R1}, an embedder-adviser framework that learns to reason over initially retrieved candidates and generate Retrieval-Centric Chain-of-Thought (\textbf{RC-CoT}). 
The adviser analyzes candidates individually to identify the discriminative cues confused by the embedder. If the target appears in the initial top-$k$ set, UniME-R1 directly reranks the candidates; otherwise, it generates RC-CoT to refine the retrieval direction and performs full-corpus re-retrieval with a dual-mode embedder.
To train the framework, we mine hard negatives to simulate realistic retrieval failures, jointly optimize direct retrieval and RC-CoT-augmented retrieval, and align the adviser with retrieval outcomes through supervised learning and retrieval-oriented reinforcement learning. Extensive experiments on MMEB-V2 and a diverse set of general multimodal retrieval benchmarks demonstrate that UniME-R1 consistently improves retrieval performance over strong baselines. 
\end{abstract}


\section{Introduction}
Unified multimodal retrieval aims to identify candidates that satisfy user intent expressed through heterogeneous inputs, including text, images, videos, and visual documents~\citep{clip,faysse2025colpali}. Such intent is often distributed implicitly across modalities and depends on fine-grained cues, such as attributes, relations, temporal events, and document-level details. Recent methods adapt Large Vision-Language Models (LVLMs)~\citep{liu2023visual,yang2025qwen3,chen2025moca,liu2025rematch} into unified encoders that map diverse inputs to a shared representation space~\citep{jiang2024vlm2vec,meng2025vlm2vecv2}. Although efficient and scalable, directly encoding the raw input into a single embedding can miss the discriminative evidence needed to distinguish the target from semantically similar candidates~\citep{xu2025mm,thirukovalluru2026breaking}.

Recent \emph{Reasoner--Embedder} methods address this limitation by using LVLMs to generate captions, query expansions, or Chain-of-Thought (CoT) rationales before embedding~\citep{lan2025ume,cui2025think,jiang2026embed}. These intermediate descriptions can enrich the query with semantics that are difficult to capture through direct encoding. However, they are typically generated from the query alone, without observing the retriever's actual outputs. Consequently, existing methods can reason about \emph{what the query describes}, but not \emph{what the retriever misunderstands}. Their CoT may restate salient content while overlooking the subtle distinction that separates the target from confusing candidates. Applying such reasoning indiscriminately can also introduce redundant or noisy semantics when the initial retrieval is already reliable.

\begin{figure}[t!]
    \centering
    \includegraphics[width=0.7\linewidth]{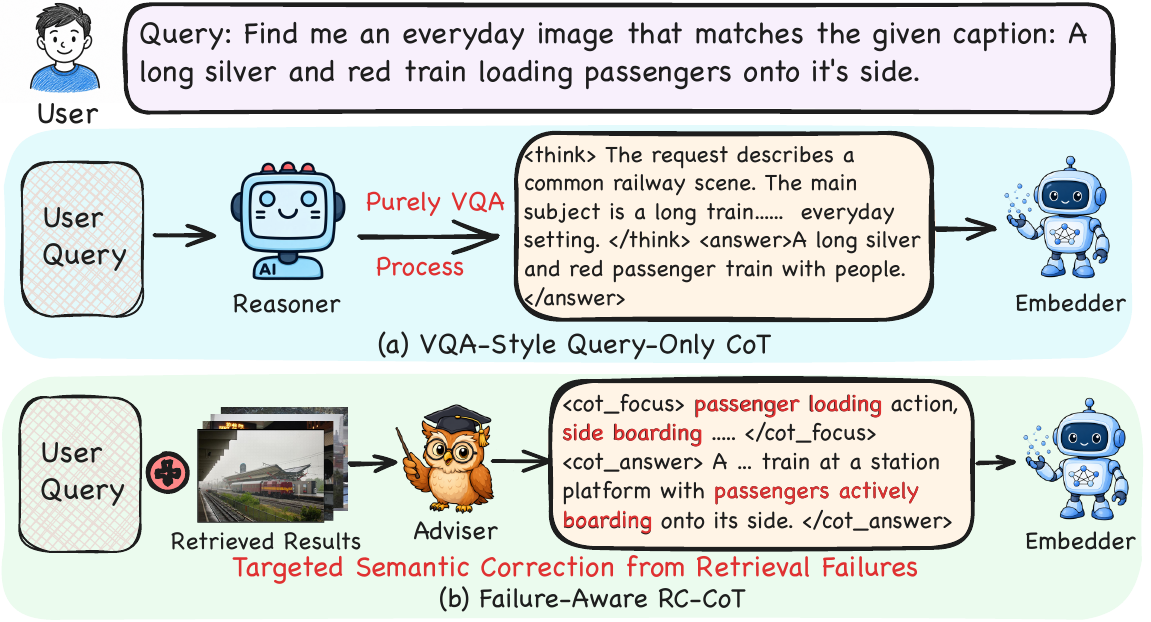}
    \vspace{-0.1in}
    \caption{\textbf{Comparison of generic CoT and RC-CoT.}
    Generic query-only CoT broadly enriches query semantics, whereas RC-CoT analyzes retrieved candidates to diagnose failures and correct the retrieval direction with targeted evidence.}
    \vspace{-0.2in}
    \label{fig:intro}
\end{figure}

We argue that effective retrieval reasoning should instead be grounded in feedback from the initial retrieval. The retrieved candidates provide model-specific evidence of the embedder's current confusion: semantically similar results reveal exactly which objects, attributes, relations, or temporal cues it fails to distinguish. For example, video candidates may contain the correct objects but present the events in the wrong temporal order. Analyzing these failures motivates our \textbf{Retrieval-Centric Chain-of-Thought (RC-CoT)}. As illustrated in Figure~\ref{fig:intro}, RC-CoT identifies the confusion exposed by retrieved candidates and emphasizes the discriminative cues required for the next retrieval. In short, \emph{generic CoT explains the query, whereas RC-CoT diagnoses and corrects a retrieval failure}.

Based on this insight, we propose \textbf{UniME-R1}, an \emph{Embedder--Adviser} framework that reasons from retrieval feedback. As shown in Figure~\ref{fig:method}, a dual-mode embedder first retrieves an initial top-$k$ candidate set, after which a retrieval-aware adviser compares the query against each candidate and identifies why the retrieved results are plausible yet insufficient. UniME-R1 explicitly separates this failure diagnosis from semantic correction. The \cotfocus field distills the attributes, relations, events, or document details that the embedder confuses, while \cotanswer converts these cues into a concise retrieval-oriented description that complements the original query. Together, they form RC-CoT, grounding query refinement in the embedder's observed failure rather than in generic query elaboration.

However, not every query requires RC-CoT-enhanced re-retrieval. When a matching candidate is already present in the initial top-$k$ set, searching the full corpus again is redundant and may introduce unnecessary semantic noise. UniME-R1 therefore adopts an adaptive rerank-or-retrieve strategy. Based on the candidate-wise analysis, the adviser predicts whether a match exists in the current candidate set and independently produces a candidate ranking. If a match is likely present, UniME-R1 directly reranks the retrieved candidates; otherwise, it appends RC-CoT to the original query, extracts a new representation through the generative embedding mode, and performs full-corpus re-retrieval. Candidates are always encoded with the discriminative embedding mode, so their representations are computed only once and remain independent of the generated reasoning. UniME-R1 can thus correct failed retrieval directions without requiring candidate-side CoT or rebuilding the candidate index, while avoiding unnecessary re-retrieval for already solvable queries.

We train UniME-R1 with hard negatives that serve complementary roles throughout the framework. They strengthen contrastive learning for the dual-mode embedder, instantiate realistic retrieval-failure contexts for adviser supervision, and support retrieval-oriented reward computation. The adviser is first trained on structured teacher annotations and then optimized with Group Relative Policy Optimization (GRPO)~\citep{deepseekmath}. Four rewards evaluate format validity, path judgment, candidate ranking, and whether the generated RC-CoT improves re-retrieval. 
Our main contributions are summarized as follows:
\begin{itemize}[leftmargin=*,noitemsep,topsep=2pt]
    \item We introduce \textbf{RC-CoT},  which diagnoses model-specific confusion from retrieved candidates and produces targeted cues for correcting retrieval.
    \item We propose \textbf{UniME-R1}, an Embedder--Adviser framework that adaptively chooses between candidate reranking and RC-CoT-enhanced full-corpus re-retrieval while retaining a reusable candidate index.
    \item We conduct extensive experiments on MMEB-V2 and diverse multimodal retrieval tasks demonstrate that UniME-R1 consistently outperforms both embedding-only and Reasoner--Embedder baselines, without requiring candidate-side CoT generation.
\end{itemize}

\section{Related Work}

\subsection{Universal Multimodal Embedding Models}
Universal Multimodal Embedding aims to learn semantically meaningful representations while simultaneously mapping diverse modalities into a shared feature space. 
As an early representative work, CLIP~\citep{clip} demonstrates strong image-text retrieval performance through large-scale cross-modal contrastive learning. However, constrained by its dual-encoder architecture, it inevitably suffers from a modality gap~\citep{gu2025unime}. 
To address this issue, recent works employ MLLMs for unified multimodal representation learning. VLM2Vec~\citep{jiang2024vlm2vec} and VLM2Vec-V2~\citep{meng2025vlm2vecv2} establish MMEB\&MMEB-V2 benchmarks for systematically evaluating embedding models. UniME-V2~\citep{gu2026unimev2} introduces an MLLM-judgment-based distribution alignment framework, while RzenEmbed~\citep{rzenembed} adopts a two-stage training strategy to improve representation discriminability. Despite these advances, existing models still struggle to distinguish hard negatives.

\subsection{Large Multimodal Reasoning Models}
Recent advances show that LVLMs benefit substantially from improved reasoning capabilities. Early methods use CoT prompting to elicit step-by-step rationales~\citep{gao2024cantor}. Inspired by DeepSeek-R1~\citep{deepseekr1}, recent studies apply reinforcement learning to optimize reasoning trajectories across diverse tasks. For instance, DeepEyes~\citep{deepeyes} relies solely on reinforcement learning to encourage thinking with images and achieves strong reasoning performance. Thyme~\citep{thyme} further extends this paradigm by enabling MLLMs to autonomously generate code for image manipulation and numerical computation, moving beyond static reasoning alone.
More recently, several studies explore the integration of multimodal reasoning into representation learning. TTE~\citep{cui2025think} employs a reasoner to generate reasoning traces that clarify complex queries and then feeds both the original query and the generated traces into an embedder. UME-R1~\citep{lan2025ume} adopts a two-stage training strategy for generative embedding, while Embed-RL~\citep{jiang2026embed} introduces an EG-RL framework in which the embedder explicitly supervises the reasoner to produce CoT traces aligned with downstream embedding objectives.
However, these approaches primarily generate CoT traces solely from the input query, which inevitably introduces irrelevant or noisy semantics due to MLLM hallucinations and the inherent ambiguity of retrieval queries. Moreover, applying such reasoning to candidate items is computationally infeasible in real-world retrieval scenarios, where the candidate pool is typically extremely large.



\begin{figure*}[t!]
    \centering
    \includegraphics[width=\linewidth]{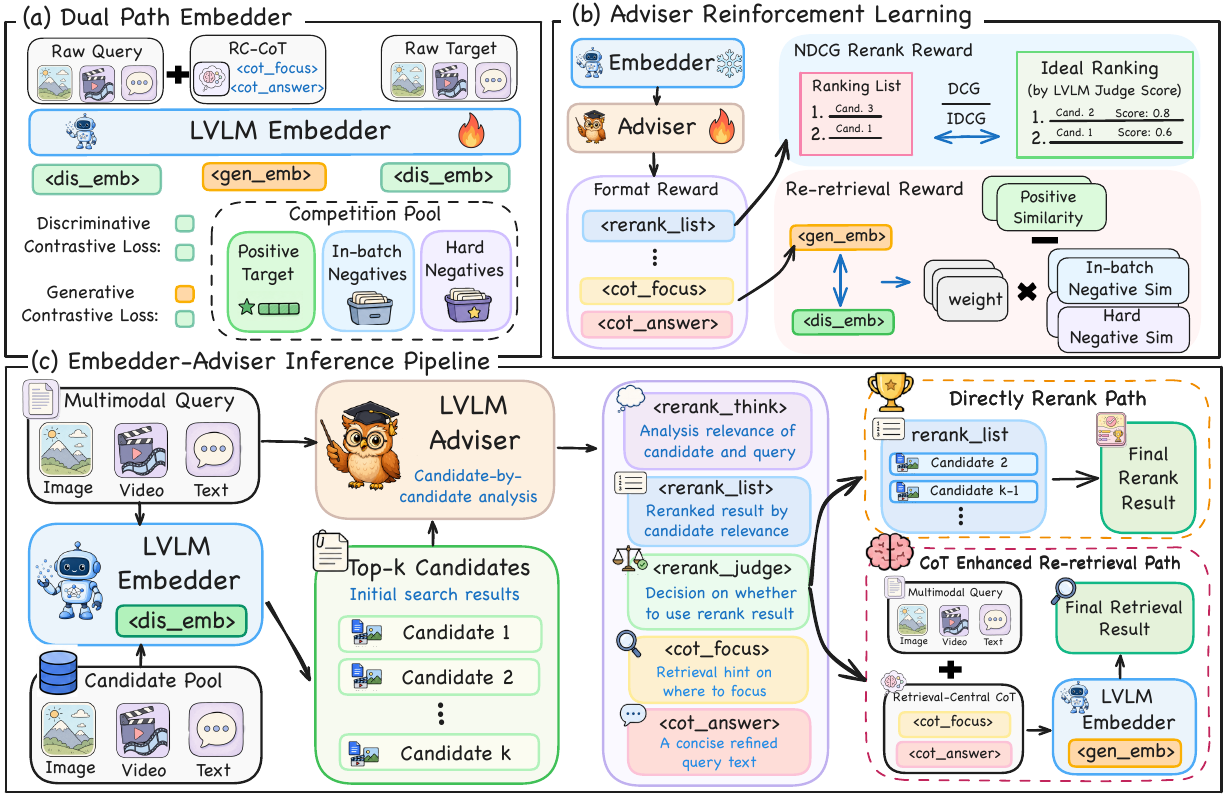}
    \vspace{-0.2in}
    \caption{\textbf{Overview of UniME-R1.} The model integrates a dual-mode embedder for direct and RC-CoT-enhanced retrieval with a retrieval-aware adviser that examines the initial top-$k$ candidates and adaptively determines whether to perform reranking or full-corpus re-retrieval with RC-CoT.}
    \vspace{-0.1in}
    \label{fig:method}
\end{figure*}

\section{UniME-R1}

\subsection{Problem Formulation}
We consider unified multimodal retrieval, where a query $q$ and each candidate $c\in\mathcal{C}$ may contain text, images, videos, or visual documents. A multimodal embedder $\Phi_{\theta}$ maps them into a shared representation space and retrieves
\begin{equation}
    \mathcal{R}_k(q)=\operatorname{TopK}_{c\in\mathcal{C}} s_{\theta}(q,c),
\end{equation}
where $s_{\theta}(q,c)$ denotes query--candidate similarity. If the target $c^{+}$ is in $\mathcal{R}_k(q)$, it can be recovered by reranking; otherwise, the query representation must be refined for full-corpus re-retrieval. We define the corresponding training label as:
\begin{equation}
    g=\mathbb{1}\!\left[c^{+}\in\mathcal{R}_k(q)\right],
\end{equation}
which is unavailable at inference time.

\subsection{Framework Overview}

As shown in Figure~\ref{fig:method}, UniME-R1 consists of two separate models: a dual-mode multimodal embedder $\Phi_{\theta}$ and a retrieval-aware adviser $\Phi_{\phi}^{\mathrm{adv}}$. The embedder first represents the raw query with \disemb\ and retrieves $\mathcal{R}_k(q)$. The adviser then takes the query and the retrieved candidates as input, producing a structured response with five designated fields: $y$ = \{\rerankthink, \reranklist, \rerankjudge, \cotfocus, \cotanswer\}. The \rerankthink\ field records candidate-by-candidate relevance analysis, \reranklist\ specifies the predicted candidate order, and \rerankjudge\ predicts whether a matching candidate exists in the current top-$k$ set. The remaining two fields form the retrieval-centric chain-of-thought (RC-CoT) used to refine the query. At inference time, the adviser adaptively routes each query to either candidate reranking or RC-CoT-enhanced re-retrieval.

\subsection{Retrieval-Centric Chain-of-Thought}

\noindent\textbf{Retrieval Failure Diagnosis.}
Generic CoT is usually generated from the query alone and therefore focuses on explaining or enriching its semantic content. In contrast, RC-CoT is conditioned on the actual retrieval results. Given $q$ and $\mathcal{R}_k(q)$, the adviser performs candidate-by-candidate analysis:
\begin{equation}
    y
    =
    \Phi_{\phi}^{\mathrm{adv}}
    \left(q,\mathcal{R}_k(q)\right).
\end{equation}
The retrieved candidates reflect the discriminative weaknesses of the current embedder. For example, they may contain the correct objects but differ in attributes, spatial relations, temporal order, or document-level details. By comparing the query with these candidates, the adviser identifies the intent-defining cues that are weakly encoded in the initial query representation.

\noindent\textbf{RC-CoT Construction.}
We construct RC-CoT by concatenating two complementary fields:
\begin{equation}
    r = [f;a],
    \qquad
    f=\cotfocus,
    \qquad
    a=\cotanswer,
\end{equation}
where $f$ is a retrieval hint that summarizes the discriminative cues exposed by the retrieved candidates, and $a$ converts this hint into a concise retrieval-oriented description. Their roles serve distinct purposes: \cotfocus\ identifies the adjustment needed for the current retrieval direction, and \cotanswer\ expresses the corrected intent in a form that can be consumed by the embedder. Thus, rather than merely adding more semantic content, RC-CoT aims to move the query representation away from the observed confusions.


\noindent\textbf{Rerank-or-Retrieve Inference.}
Applying RC-CoT-enhanced re-retrieval to every query is neither necessary nor efficient. When the target is already ranked near the top of the initial results, the original query representation typically captures the user intent sufficiently, and further refinement may introduce redundant or noisy semantics. In such cases, reranking alone is usually adequate to identify the positive candidate. By contrast, when the initial retrieval fails to include a matching candidate, reranking cannot recover the target, and RC-CoT becomes necessary to redirect retrieval toward more relevant results. This adaptive strategy also reduces computation by avoiding an additional search over the full candidate pool when reranking alone is sufficient.

Let $\hat{g}\in\{0,1\}$ denote the binary decision represented by \rerankjudge. If $\hat{g}=1$, UniME-R1 returns the candidates in the order specified by \reranklist. If $\hat{g}=0$, it constructs the refined query $[q;r]$ and performs full-corpus re-retrieval:
\begin{equation}
\begin{aligned}
\mathcal{R}^{\,r}_k(q)
= \operatorname{TopK}_{c \in \mathcal{C}} \;
s\bigl(&\Phi_{\theta}([q;r], \genemb), \Phi_{\theta}(c, \disemb)\bigr).
\end{aligned}
\end{equation}
The path decision and candidate ranking are represented by separate fields: \rerankjudge\ only determines whether a match exists, whereas \reranklist\ determines the order of the retrieved candidates.

\begin{figure*}[t!]
    \centering
    \includegraphics[width=\linewidth]{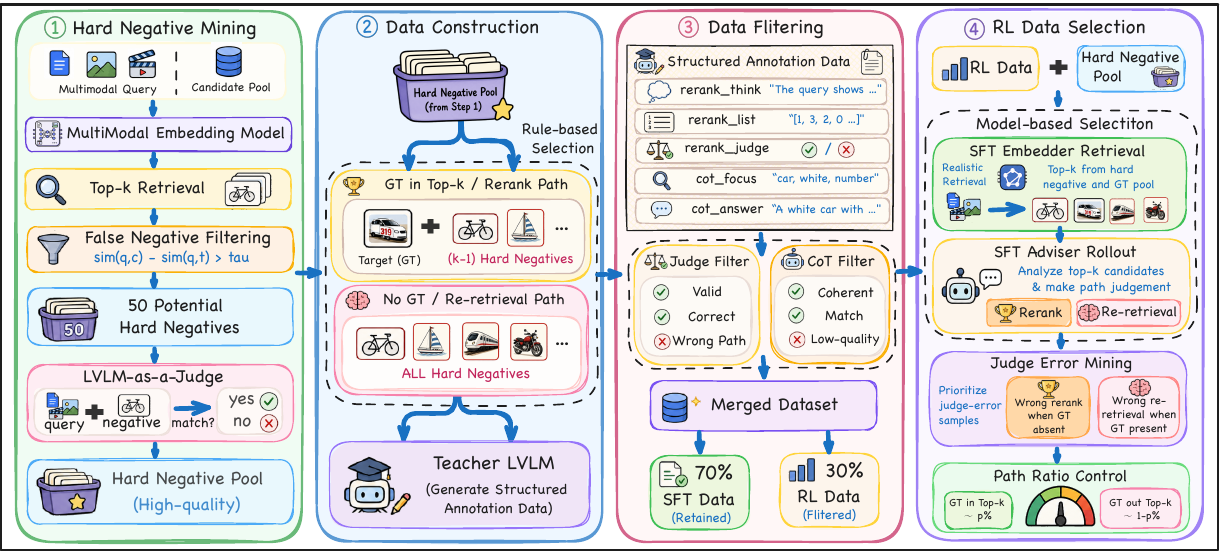}
    \vspace{-0.2in}
    \caption{\textbf{Training data construction for UniME-R1.} Mined hard negatives are used to construct target-in and target-out candidate sets for supervised learning, and model-retrieved candidate sets aligned with the SFT embedder for GRPO.}
    \vspace{-0.1in}
    \label{fig:data_pipeline}
\end{figure*}

\subsection{Dual-Mode Multimodal Embedder}

\noindent\textbf{Representation Design.}
As illustrated in Figure~\ref{fig:method}(a), the embedder supports both initial retrieval and RC-CoT-enhanced re-retrieval by constructing two dedicated query representations. Given a query $q$ and its RC-CoT $r$, we define these representations as follows:
\begin{equation}
    \mathbf{z}^{d}_{q}
    =
    \Phi_{\theta}(q,\disemb),
    \qquad
    \mathbf{z}^{g}_{q}
    =
    \Phi_{\theta}([q;r],\genemb).
\end{equation}
The discriminative representation $\mathbf{z}^{d}_{q}$ is used for initial retrieval, while the RC-CoT-enhanced representation $\mathbf{z}^{g}_{q}$ is used for re-retrieval. Candidates are always encoded through \disemb:
\begin{equation}
    \mathbf{z}^{d}_{c}
    =
    \Phi_{\theta}(c,\disemb).
\end{equation}
Consequently, candidate representations remain independent of the adviser output and can be precomputed and indexed once. Both query modes search the same candidate index.

\noindent\textbf{Joint Contrastive Training.}
For each training query $q_i$, we construct a candidate pool $\mathcal{P}_i$ containing the target $c_i^{+}$, in-batch negatives, and $m$ mined hard negatives. Given a query representation $\mathbf{z}_q$, we define:
\begin{equation}
    \ell(\mathbf{z}_{q},c_i^{+};\mathcal{P}_i)
    =
    -\log
    \frac{
    \exp(s(\mathbf{z}_{q},\mathbf{z}^{d}_{c_i^{+}})/\tau)
    }{
    \sum_{c\in\mathcal{P}_i}
    \exp(s(\mathbf{z}_{q},\mathbf{z}^{d}_{c})/\tau)
    },
\end{equation}
where $s(\cdot,\cdot)$ is cosine similarity and $\tau$ is the temperature. The raw and RC-CoT-enhanced query representations are jointly optimized:
\begin{equation}
    \mathcal{L}_{\mathrm{dis}}
    =
    \ell(\mathbf{z}^{d}_{q_i},c_i^{+};\mathcal{P}_i),
    \qquad
    \mathcal{L}_{\mathrm{gen}}
    =
    \ell(\mathbf{z}^{g}_{q_i},c_i^{+};\mathcal{P}_i),
\end{equation}
    
The RC-CoT used by $\mathcal{L}_{\mathrm{gen}}$ is generated by the annotation teacher. Through the shared hard-negative-enriched pool, \disemb\ learns robust direct retrieval, while \genemb\ learns to translate retrieval-centric reasoning into an effective correction of the query representation. The final loss is $\mathcal{L}_{\mathrm{emb}}=\mathcal{L}_{\mathrm{dis}} + \mathcal{L}_{\mathrm{gen}}.$

\subsection{Retrieval-Aware Adviser}
\noindent\textbf{Supervised Fine-Tuning.}
The adviser is initially trained to generate the complete structured output using a standard autoregressive objective:
\begin{equation}
    \mathcal{L}_{\mathrm{adv}}^{\mathrm{SFT}}
    =
    -\sum_{t=1}^{|y|}
    \log
    p_{\phi}
    \left(y_t
    \mid
    y_{<t},q,\mathcal{S}(q)
    \right),
\end{equation}
where $\mathcal{S}(q)$ is either $\mathcal{S}^{\mathrm{in}}(q)$ or $\mathcal{S}^{\mathrm{out}}(q)$.

\noindent\textbf{GRPO with Retrieval-Oriented Rewards.}
As shown in Figure~\ref{fig:method}(b), after SFT, we freeze the embedder and optimize only the adviser using GRPO. We define four complementary reward components:
\begin{equation}
    R
    =
    \lambda_f R_{\mathrm{fmt}}
    +
    \lambda_j R_{\mathrm{judge}}
    +
    \lambda_r R_{\mathrm{rank}}
    +
    \lambda_c R_{\mathrm{cot}}.
\end{equation}
Because every rollout contains all five fields, all four rewards are computed for every sample.

\noindent\textit{Format Reward.}
The format reward $R_{\mathrm{fmt}}$ evaluates whether all required XML-style fields are present and well-formed. It further verifies that \reranklist\ forms a valid permutation of candidate indices in the current top-$k$ set. 

\noindent\textit{Judge Reward.}
The judge reward directly evaluates the binary path decision:
\begin{equation}
    R_{\mathrm{judge}}
    =
    \mathbb{1}[\hat{g}=g],
\end{equation}
where $\hat{g}$ is predicted by \rerankjudge\ and $g$ indicates whether the target occurs in the candidate set.

\noindent\textit{Rerank Reward.}
We reuse the LVLM-judge relevance scores from hard negative mining as supervision for candidate ordering. Let $\mathbf{a}=[a_1,\ldots,a_k]$ denote the scores of the current candidates and let $\pi$ be the permutation predicted by \reranklist. We then define:
\begin{equation}
    R_{\mathrm{rank}}
    =
    \operatorname{NDCG}(\pi;\mathbf{a})
    =
    \frac{
    \operatorname{DCG}(\pi;\mathbf{a})
    }{
    \operatorname{IDCG}(\mathbf{a})
    }.
\end{equation}
This reward is computed for both target-in and target-out candidate sets, encouraging the adviser to produce a coherent relevance ranking under both inference branches.

\noindent\textit{RC-CoT Reward.}
We concatenate \cotfocus\ and \cotanswer\ to obtain $r$ and encode $[q;r]$ with the frozen embedder through \genemb. The reward then evaluates whether retrieval based on the resulting representation ranks the positive candidate above the hard negatives:
\begin{equation}
    R_{\mathrm{cot}}
    =
    \operatorname{MRR}
    \left(c^{+};\{c^{+}\}\cup\mathcal{N}\right)
    \cdot
    \Delta(q,r),
\end{equation}
where $\Delta(q,r)=s(\mathbf{z}^{g}_{q},\mathbf{z}^{d}_{c^{+}})-\bar{s}(\mathbf{z}^{g}_{q},\mathbf{z}^{d}_{\mathcal{N}}).$

The evaluation pool contains the target and a negative set $\mathcal{N}$ composed of in-batch negatives and mined hard negatives. The first term rewards a high target rank, while the similarity margin rewards RC-CoT that moves the refined query toward the target and away from confusing candidates. 



\subsection{Training Data Construction}

\noindent\textbf{Hard Negative Mining.}
Hard negatives serve two roles in UniME-R1: strengthening contrastive learning and creating realistic retrieval-failure contexts for adviser training. As shown in Figure~\ref{fig:data_pipeline}, following~\cite{gu2026unimev2}, our mining pipeline consists of three stages: (1) retrieving top-ranked candidates with an off-the-shelf multimodal embedder, (2) filtering potential false negatives based on similarity gaps, and (3) verifying candidate relevance using an LVLM as a judge. Further details are provided in the Appendix~\ref{data_statistic}.

\noindent\textbf{Rule-Based SFT Data.}
We construct two candidate-set types to cover both inference paths. To simulate target-in-top-$k$, we combine the target with $k-1$ hard negatives:
\begin{equation}
    \mathcal{S}^{\mathrm{in}}(q)
    =
    \{c^{+}\}
    \cup
    \mathcal{H}_{k-1}(q).
\end{equation}
To simulate target-out-of-top-$k$, we use only hard negatives:
\begin{equation}
    \mathcal{S}^{\mathrm{out}}(q)
    =
    \mathcal{H}_{k}(q).
\end{equation}
For each instance, a strong LVLM teacher observes only the query and the constructed candidate set and generates all five structured fields. 

We filter the annotations in two steps. A decision filter verifies whether \rerankjudge\ selects the correct path. An RC-CoT filter uses an LVLM to evaluate whether \cotfocus\ and \cotanswer\ are coherent and retrieval-relevant. The retained examples form supervised data for both the adviser and the embedder’s RC-CoT branch.

\noindent\textbf{Embedder-Aligned GRPO Data.}
Rule-based candidate sets provide controlled supervision but do not fully match the distribution encountered after SFT. We therefore construct a second dataset aligned with the current models. For each query, the SFT embedder retrieves top-$k$ candidates from a pool consisting of the positive candidate and its mined hard negatives, and the SFT adviser performs 8 rollouts on the retrieved set. Based on the \rerankjudge\ field, we discard samples for which all eight rollouts are either correct or incorrect, as they are too easy or too difficult to provide informative relative supervision. We additionally balance target-in and target-out instances to avoid bias toward either inference path. This construction focuses GRPO on the observed failure modes of the SFT system while preserving realistic, embedder-specific distractors.


\begin{table*}[t]
\centering
\caption{Comparison of baseline methods on MMEB-V2. Given the diversity of model backbones, we aggregate results by model size. Models with 2B--3B parameters are categorized as \emph{small}, while those with 4B--7B parameters are categorized as \emph{medium}. $^\dagger$ indicates that for the TTE model, we adopt the student variant to ensure a fair comparison without relying on a large external teacher model. Metrics are abbreviated as follows: \textbf{CLS} (classification), \textbf{QA} (question answering), \textbf{RET} (retrieval), \textbf{GD} (grounding), \textbf{MRET} (moment retrieval), \textbf{VDR} (ViDoRe), \textbf{VR} (VisRAG), and \textbf{OOD} (out-of-domain). The best and second-best results are \textbf{bolded} and \underline{underlined}.}
\label{tab:main_result}
\vspace{-0.05in}

\renewcommand{\arraystretch}{1} 
\tabcolsep2pt
\resizebox{\linewidth}{!}{
\begin{tikzpicture}
\node[inner sep=2.5pt] (tbl) {
  {    
    \begin{NiceTabular}{llccccc ccccc ccccc c}

\multirow{2}{*}{\textbf{Model}} 
& \multirow{2}{*}{\textbf{Backbone}} 
& \multicolumn{5}{c}{\textbf{Image}} 
& \multicolumn{5}{c}{\textbf{Video}} 
& \multicolumn{5}{c}{\textbf{VisDoc}}
& \multirow{2}{*}{\textbf{All}}\\
\cmidrule(lr){3-7} \cmidrule(lr){8-12} \cmidrule(lr){13-17}
& &\textbf{CLS} & \textbf{QA} & \textbf{RET} & \textbf{GD} & \textbf{Overall} 
& \textbf{CLS} & \textbf{QA} & \textbf{RET} & \textbf{MRET} & \textbf{Overall} 
& \textbf{VDRv1} & \textbf{VDRv2} & \textbf{VR} & \textbf{OOD} & \textbf{Overall} \\
\midrule

\multicolumn{18}{c}{\textbf{\textit{Small-size Models}}} \\
\midrule

GME & Qwen2-VL-2B & 
54.4 & 29.9 & 66.9 & 55.5 & 51.9 & 
34.9 & 42.0 & 25.6 & 32.4 & 33.9 & 
\underline{86.1} & \underline{54.0} & 82.5 & 43.1 & 72.7 & 
54.1 \\

ColPali-V1.3 & PaliGemma-3B & 
40.3 & 11.5 & 48.1 & 40.3 & 34.9 & 
26.7 & 37.8 & 21.6 & 25.5 & 28.2 & 
83.6 & 52.0 & 81.1 & 43.1 & 71.0 & 
44.4 \\

VLM2Vec & Qwen2-VL-2B & 
58.7 & 49.3 & 65.0 & 72.9 & 59.7 & 
33.4 & 30.5 & 20.6 & 33.0 & 29.0 & 
49.8 & 13.5 & 51.8 & 33.5 & 41.6 & 
47.0 \\

RzenEmbed-V1 & Qwen2-VL-2B&
65.3 & 61.7 & \textbf{73.8} & 77.9 & 68.5 & 
45.6 & 47.5 & 38.3 & 36.7 & 42.6 &
\textbf{87.0} & \textbf{57.6} & \underline{85.4} & 43.3 & \underline{74.4} &
64.4 \\

VLM2Vec-V2 & Qwen2-VL-2B & 
62.9 & 56.3 & 69.5 & 77.3 & 64.9 &
39.3 & 34.3 & 28.8 & 38.5 & 34.9 & 
75.5 & 44.9 & 79.4 & 39.4 & 65.4 & 
58.0 \\

UME-R1 & Qwen2-VL-2B & 
64.8 & 62.8 & 67.6 & 77.2 & 66.6 &
44.3 & 51.2 & 32.9 & 39.7 & 42.2 & 
72.4 & 46.2 & 79.2 & 37.2 & 63.9 & 
60.1 \\

$\text{TTE}_s^{\dagger}$ &Qwen3-VL-2B &
\textbf{67.9} & 66.6 & 70.2 & 84.1 & \underline{70.1} &
47.3 & 49.1 & 34.4 & 33.2 & 32.1 &
77.5 & 53.2 & 83.2 & 41.1 & 68.8 &
63.1 \\

Embed-RL & Qwen3-VL-2B & 
62.8 & \underline{67.9} & 68.6 & \textbf{90.4} & 69.2 & 
\textbf{57.0} & \underline{55.9} & \textbf{45.1} & \underline{49.4} & \underline{52.1} & 
79.9 & 52.0 & 84.6 & \underline{65.7} & 74.1 & 
\underline{66.8} \\ 

\hdashline
\textbf{UniME-R1} & Qwen3-VL-2B & 
\underline{67.2} & \textbf{73.5} & \underline{73.4} & \underline{88.1} & \textbf{73.4} & 
\underline{56.0} & \textbf{66.5} & \underline{41.8} & \textbf{49.8} & \textbf{53.9} & 
83.4 & 53.1 & \textbf{87.1} & \textbf{67.6} & \textbf{76.6} & 
\textbf{69.9} \\ 

\midrule

\multicolumn{18}{c}{\textbf{\textit{Medium-size Models}}} \\

\midrule

GME & Qwen2-VL-7B & 
57.7 & 34.7 & 71.2 & 59.3 & 56.0 & 
37.4 & 50.4 & 28.4 & 38.2 & 38.6 & 
\underline{89.4} & 55.6 & 85.0 & 44.4 & 75.2 & 
57.8 \\


LamRA & Qwen2.5-VL-7B & 
51.7 & 34.1 & 66.9 & 56.7 & 52.4 & 
32.9 & 42.6 & 23.2 & 37.6 & 33.7 & 
56.3 & 33.3 & 58.2 & 40.1 & 50.2 & 
47.4 \\

VLM2Vec & Qwen2-VL-7B & 
62.7 & 56.9 & 69.4 & 82.2 & 65.5 & 
39.1 & 30.0 & 29.0 & 40.6 & 34.0 & 
56.9 & 9.4 & 59.1 & 38.1 & 46.4 & 
52.3 \\

CAFe & LLaVA-OV-7B & 
63.6 & 61.7 & 69.1 & 87.6 & 67.6 & 
35.8 & 58.7 & 34.4 & 39.5 & 42.4 & 
70.7 & 49.6 & 79.5 & 38.1 & 63.9 & 
60.6 \\

RzenEmbed-V1 &  Qwen2-VL-7B &
\textbf{69.8} & 68.7 & \textbf{76.8} & 85.7 & 73.6 &
52.8 & 56.2 & 41.9 & 41.8 & \underline{48.9} &
\textbf{89.5} & \underline{60.8} & \underline{87.9} & 44.4 & \underline{76.8} &
\underline{68.9} \\

UME-R1 & Qwen2-VL-7B & 
67.1 & 69.2 & 71.9 & 84.9 & 71.3 &
48.6 & \underline{60.7} & 38.2 & 39.3 & 47.5 & 
75.7 & 50.5 & 83.7 & 37.6 & 67.1 & 
64.5 \\

$\text{TTE}_s^{\dagger}$ & Qwen2-VL-7B &
\underline{69.7} & \underline{72.4} & 74.0 & 90.6 & \underline{74.2} &
49.1 & 60.6 & 36.4 & 37.2 & 46.8 &
84.1 & \textbf{62.7} &\textbf{ 91.9} & 47.6 & 76.4 &
68.6 \\

Embed-RL & Qwen3-VL-4B & 
63.7 & 70.5 & 71.3 & \underline{91.4} & 70.1 & 
\underline{57.6} & 58.4 & \underline{45.1} & \underline{49.5} & \textbf{53.0} & 
80.2 & 53.4 & 84.9 & \textbf{67.1} & 74.7 & 
68.1 \\ 

\hdashline
\textbf{UniME-R1} & Qwen3-VL-4B & 
67.7 & \textbf{74.2} & \underline{74.5} & \textbf{94.9} & \textbf{74.3} & 
\textbf{60.7} & \textbf{61.1} & \textbf{51.6} & \textbf{50.6} & \textbf{53.0} & 
83.0 & 60.6 & 83.7 & \underline{66.6} & \textbf{77.4} & 
\textbf{70.3} \\ 

    \end{NiceTabular}
  }%
};
\draw[line width=0.08pt, rounded corners=4pt]
  (tbl.south west) rectangle (tbl.north east);
\draw[line width=0.08pt, rounded corners=4pt, opacity=0.6]
  ([xshift=0.5pt,yshift=0.5pt]tbl.south west)
    rectangle
  ([xshift=0.5pt,yshift=0.5pt]tbl.north east);
\end{tikzpicture}
}
\vspace{-0.1in}
\end{table*}

\section{Experiment}

\subsection{Implementation Details}
We train Qwen3-VL-2B~\citep{yang2025qwen3} and Qwen3-VL-4B~\citep{yang2025qwen3} as Embedders and adopt a sub-batch strategy following
VLM2Vec~\citep{jiang2024vlm2vec}. The models are trained for 2 epochs with an initial learning rate of
1e-4 and weight decay of 0.01; the batch size is set to 512, and we use Low-Rank Adaptation (LoRA)~\citep{hu2022lora} for fine-tuning. For Adviser, we train Qwen3-VL-4B~\citep{yang2025qwen3} with LlamaFactory for SFT with the batch size of 512. For 
reinforcement learning, we train Adviser with GRPO
algorithm~\citep{guo2025deepseek} for 1 epoch, with a batch size of 256, learning rate of 3e-6, and standard GRPO hyperparameters.

\subsection{Datasets and Evaluation}
All UniME-R1 training data (for both the embedder and adviser) are derived from the MMEB-V2 training set~\citep{meng2025vlm2vecv2}. For the dual-mode embedder, \disemb is trained on the original MMEB-V2 data, while \genemb is trained on an additional 1.73M RC-CoT-augmented samples constructed from MMEB-V2. For the retrieval-aware adviser, we construct 643K structured samples for SFT and 13K examples for GRPO training. To further evaluate UniME-R1’s general multimodal retrieval capability, we conduct experiments on diverse cross-modal benchmarks, including short-caption image--text retrieval on Flickr30K~\cite{flickr30k} and COCO2014~\cite{MSCOCO2014}, long-caption image--text retrieval on ShareGPT4V~\cite{sharegpt4v} and Urban1K~\cite{longclip}, and video retrieval on UVRB~\cite{guo2025towards}. Detailed data statistics and downstream results are provided in the Table~\ref{tab:adviser_data_stats}.

\subsection{Baselines}
We compare UniME-R1 with two categories of representative baselines. \textbf{Embedder-only methods:} VLM2Vec~\citep{jiang2024vlm2vec}, VLM2Vec-V2~\citep{meng2025vlm2vecv2}, GME~\citep{zhang2024gme}, ColPali~\citep{faysse2025colpali}, CAFe~\citep{yu2025cafe}, B3~\citep{thirukovalluru2026breaking}, Unite~\citep{kong2025modality}, LamRA~\citep{liu2025lamra}, and RzenEmbed-v1~\citep{rzenembed}. \textbf{Reasoner--Embedder methods:} UME-R1~\citep{lan2025ume}, TTE~\citep{cui2025think}, and Embed-RL~\citep{jiang2026embed}.
\begin{table*}[t]
\centering
\caption{Zero-shot performance comparison on general multimodal retrieval tasks. We report Recall@1 on Flickr30K, COCO, ShareGPT4V, and Urban1K, and average performance on UVRB. The best and second-best results are \textbf{bolded} and \underline{underlined}.}
\label{tab:zero-shot-retrieval}
\vspace{-0.05in}

\renewcommand{\arraystretch}{1}
\tabcolsep2pt
\resizebox{0.9\linewidth}{!}{
\begin{tikzpicture}
\node[inner sep=2.5pt] (tbl) {
{
\begin{NiceTabular}{lcccccccccc}
\multirow{2}{*}{\textbf{Model}} 
& \multirow{2}{*}{\textbf{Backbone}} 
& \multicolumn{2}{c}{\textbf{Flickr30K}} 
& \multicolumn{2}{c}{\textbf{COCO}}
& \multicolumn{2}{c}{\textbf{ShareGPT4V}} 
& \multicolumn{2}{c}{\textbf{Urban1K}}
& \multicolumn{1}{c}{\textbf{UVRB}} \\
\cmidrule(lr){3-4} \cmidrule(lr){5-6} \cmidrule(lr){7-8} \cmidrule(lr){9-10} \cmidrule(lr){11-11}
& &  $\mathbf{q^i\!\rightarrow\!c^t}$ & $\mathbf{q^t\!\rightarrow\!c^i}$ 
& $\mathbf{q^i\!\rightarrow\!c^t}$ & $\mathbf{q^t\!\rightarrow\!c^i}$ 
& $\mathbf{q^i\!\rightarrow\!c^t}$ & $\mathbf{q^t\!\rightarrow\!c^i}$ 
& $\mathbf{q^i\!\rightarrow\!c^t}$ & $\mathbf{q^t\!\rightarrow\!c^i}$
& \textbf{AVG}  \\
\midrule
E5-V &LLaVA-1.6-7B  &85.7  &77.3  &57.6  &49.1 &82.1  &85.1 &83.2 &88.9 &-  \\
UniME-V2 &Qwen2-VL-7B  &93.5  &84.6  &70.3 &57.3 &95.2 &94.3 &96.3 &97.2 &	49.3  \\
B3 &Qwen2-VL-7B  &95.9  &85.5  &\underline{77.6}  &\textbf{62.8} &\underline{98.0}  &\textbf{98.1} &-- &-- &53.8  \\
Unite &Qwen2-VL-7B  &94.4  &\underline{86.1}  &--  &-- &93.2  &93.3 &95.6 &95.5 &55.9  \\
Embed-RL &Qwen3-VL-4B  &--  &--  &--  &-- &--  &-- &-- &-- &60.0  \\
\hdashline
\textbf{UniME-R1}  &Qwen3-VL-2B &\textbf{96.6}  &85.7  &\underline{77.6}  &60.5  &97.9 &97.1  &\underline{98.6} &\underline{98.8} &\underline{61.1}  \\
\textbf{UniME-R1} &Qwen3-VL-4B  &\underline{96.3}  &\textbf{86.9}  &\textbf{78.9}  &\underline{62.0} &\textbf{98.7}  &\underline{97.5} &\textbf{99.1} &\textbf{99.3} &\textbf{61.6}  \\
\end{NiceTabular}
}%
};
\draw[line width=0.08pt, rounded corners=4pt]
  (tbl.south west) rectangle (tbl.north east);
\draw[line width=0.08pt, rounded corners=4pt, opacity=0.6]
  ([xshift=0.5pt,yshift=0.5pt]tbl.south west)
    rectangle
  ([xshift=0.5pt,yshift=0.5pt]tbl.north east);
\end{tikzpicture}
}
\vspace{-0.1in}
\end{table*}
\begin{table}[t]
\centering
\caption{Effect of retrieval feedback on CoT-enhanced retrieval. We report the aggregated results on image, video, and visual-document tasks in MMEB-V2.}
\label{tab:ablation_rc_cot}
\vspace{-0.05in}
\renewcommand{\arraystretch}{1}
\tabcolsep2pt
\resizebox{0.8\linewidth}{!}{
\begin{tikzpicture}
\node[inner sep=2.5pt] (tbl) {
{
\begin{NiceTabular}{lcc cccc}
\multirow{2}{*}{\textbf{Variant}} &
\multicolumn{2}{c}{\textbf{Reasoning Input}} &
\multicolumn{4}{c}{\textbf{MMEB-V2}} \\
\cmidrule(lr){2-3} \cmidrule(lr){4-7}
& \textbf{Context} & \textbf{\cotfocus} &
\textbf{Image} & \textbf{Video} & \textbf{VisDoc} & \textbf{Overall} \\
\midrule
Query-only CoT & Query & \ding{52} & 70.2 & 47.2 & 72.5 & 65.6 \\
Random-candidate CoT & Random & \ding{52} & 71.4 & 47.6 & 73.2 & 66.5 \\
RC-CoT w/o focus & Retrieved & \ding{56}  & 72.3 & 49.2 &  75.1 & 67.9 \\
\textbf{RC-CoT} & Retrieved & \ding{52} & \textbf{72.8} & \textbf{49.8} & \textbf{75.8} & \textbf{68.5} \\
\end{NiceTabular}
}%
};
\draw[line width=0.08pt, rounded corners=4pt]
  (tbl.south west) rectangle (tbl.north east);
\draw[line width=0.08pt, rounded corners=4pt, opacity=0.6]
  ([xshift=0.5pt,yshift=0.5pt]tbl.south west)
  rectangle
  ([xshift=0.5pt,yshift=0.5pt]tbl.north east);
\end{tikzpicture}
}
\vspace{-0.2in}
\end{table}

\subsection{Main Result}

Table~\ref{tab:main_result} compares UniME-R1 with representative methods on MMEB-V2. UniME-R1 achieves the best overall performance at both model scales, reaching 69.9 (2B) and 70.3 (4B), which surpass the strongest baseline in each group by 3.1 and 1.4 points, respectively. Compared with the most direct Reasoner--Embedder baselines (UME-R1 and TTE), the 2B model improves the overall score by 9.8 and 6.8 points, while the 4B model yields gains of 5.8 and 1.7 points. UniME-R1 also consistently outperforms embedding-only methods without CoT, such as VLM2Vec and GME, demonstrating the benefit of retrieval-centric reasoning beyond direct multimodal encoding. The improvements span image, video, and visual-document tasks, supporting the generality of retrieval-feedback RC-CoT. Finally, scaling from 2B to 4B further raises the overall score from 69.9 to 70.3; notably, the 2B model already outperforms all medium-size baselines, indicating that the gains primarily stem from the proposed framework rather than model scale alone.

We further evaluate UniME-R1 on general multimodal retrieval tasks.  As shown in Table~\ref{tab:zero-shot-retrieval}, UniME-R1-2B improves over the UniME-V2-7B by 3.1\&1.1, 7.3\&3.2, 2.7\&2.8, and 2.3\&1.6 points on Flickr30K, COCO, ShareGPT4V, and Urban1K. Similarity, UniME-R1-4B also achieves 1.6 points improvement on UVRB compared with Embed-RL-4B. The consistent gains on different general multimodal retrieval tasks indicate that retrieval-feedback RC-CoT can exploit both concise visual cues and richer descriptions.

\section{Ablation Study}

\subsection{Effectiveness of Retrieval-Centric CoT}
In the early stage of our experiments, we investigated the impact of different Retrieval-Centric CoT formats under a fixed adviser (Qwen3-VL-235B), generation budget, and dual-mode embedder. As shown in Table~\ref{tab:ablation_rc_cot}, we make three observations:
(1) \textbf{Actual retrieval feedback is essential.} Query-only CoT obtains 65.6 overall, while conditioning on random candidates provides only a modest gain to 66.5. RC-CoT based on the actual top-$k$ results reaches 68.5, improving over these variants by 2.9 and 2.0 points, respectively.
(2) \textbf{The benefit is consistent across modalities.} Compared with query-only CoT, RC-CoT improves image, video, and visual-document retrieval by 2.6, 2.6, and 3.3 points, demonstrating that failure-aware reasoning generalizes across retrieval scenarios.
(3) \textbf{Explicit failure diagnosis contributes beyond query refinement.} Removing \cotfocus\ lowers the overall score from 68.5 to 67.9, confirming that summarizing the confusion exposed by retrieved candidates provides useful guidance beyond the final refined description.

\subsection{Effect of the Rerank-or-Retrieve Strategy}
Table~\ref{tab:ablation_routing} compares fixed and adaptive inference strategies. We make three observations:
    (1) \textbf{Both paths are effective but complementary.} Always reranking and always re-retrieving improve the overall score from 63.5 to 69.1 and 69.0, respectively. Reranking recovers targets already in the top-$k$, whereas re-retrieval corrects failed retrieval directions.
    (2) \textbf{Adaptive routing performs best among learned strategies.} UniME-R1 reaches 69.9 overall, outperforming the two fixed strategies by 0.8 and 0.9 points and improving initial retrieval by 6.4 points overall and 13.0 points on video tasks. The selective-rerank row only evaluates queries routed to reranking and is therefore not directly comparable to full-set results.
    (3) \textbf{Routing still has room for improvement.} To further explore the upper bound of performance, we report the results of oracle routing, which denotes the setting in which the better result between reranking and re-retrieval is selected. Oracle routing reaches 72.2 overall, 2.3 points above learned routing, with the largest gap of 3.2 points on video tasks. Because it changes only the routing decision, it provides an upper bound on path selection rather than on reranking or RC-CoT generation.

\begin{table*}[t]
\centering
\caption{Ablation of the rerank-or-retrieve strategy. We report the aggregated results on image, video, and visual-document tasks in MMEB-V2. Selective rerank evaluates only the queries that the adviser routes to the reranking path and excludes queries assigned to RC-CoT re-retrieval.}
\label{tab:ablation_routing}
\vspace{-0.05in}
\renewcommand{\arraystretch}{1}
\tabcolsep2pt
\resizebox{0.85\linewidth}{!}{
\begin{tikzpicture}
\node[inner sep=2.5pt] (tbl) {
{
\begin{NiceTabular}{lccc cccc}
\multirow{2}{*}{\textbf{Strategy}} &
\multicolumn{3}{c}{\textbf{Path Configuration}} &
\multicolumn{4}{c}{\textbf{MMEB-V2}} \\
\cmidrule(lr){2-4} \cmidrule(lr){5-8}
& \textbf{Rerank} & \textbf{Re-retrieve} & \textbf{Routing} &
\textbf{Image} & \textbf{Video} & \textbf{VisDoc} & \textbf{Overall} \\
\midrule
Init retrieval & \ding{56} & \ding{56} & \ding{56} & 69.6 & 40.9 & 71.3 & 63.5 \\
Selective rerank & \ding{52} & \ding{56} & Adviser & 72.4 & 51.9 & 75.7 & 68.9 \\
\textbf{UniME-R1} & \ding{52} & \ding{52} & Adviser & \textbf{73.4} & \textbf{53.9} & \textbf{76.6} & \textbf{69.9} \\
\midrule
Always rerank & \ding{52} & \ding{56} & Fixed & 72.6 & 53.4 & 76.0 & 69.1 \\
Always re-retrieve & \ding{56} & \ding{52} & Fixed & 73.0 & 52.2 & 76.0 & 69.0 \\
Oracle routing & \ding{52} & \ding{52} & Oracle & \textbf{75.8} & \textbf{57.1} & \textbf{78.3} & \textbf{72.2} \\
\end{NiceTabular}
}%
};
\draw[line width=0.08pt, rounded corners=4pt]
  (tbl.south west) rectangle (tbl.north east);
\draw[line width=0.08pt, rounded corners=4pt, opacity=0.6]
  ([xshift=0.5pt,yshift=0.5pt]tbl.south west)
  rectangle
  ([xshift=0.5pt,yshift=0.5pt]tbl.north east);
\end{tikzpicture}
}
\vspace{-0.1in}
\end{table*}
\begin{table}[t]
\centering
\caption{Ablation of the dual-mode embedder. We report the aggregated results on image, video, and visual-document tasks in MMEB-V2.}
\label{tab:ablation_embedder}
\vspace{-0.05in}
\renewcommand{\arraystretch}{1}
\tabcolsep2pt
\resizebox{0.6\linewidth}{!}{
\begin{tikzpicture}
\node[inner sep=2.5pt] (tbl) {
{
\begin{NiceTabular}{lc cccc}
\multirow{2}{*}{\textbf{Objective}} &
\multirow{2}{*}{\textbf{Hard Neg.}} &
\multicolumn{4}{c}{\textbf{MMEB-V2}} \\
\cmidrule(lr){3-6}
& & \textbf{Image} & \textbf{Video} & \textbf{VisDoc} & \textbf{Overall} \\
\midrule
$\mathcal{L}_{\mathrm{dis}}$ & \ding{56} & 68.7 & 39.6 & 69.4 & 62.3 \\
$\mathcal{L}_{\mathrm{dis}}+\mathcal{L}_{\mathrm{gen}}$ & \ding{56} & 68.6 & 40.3 & 70.3 & 62.6 \\
$\mathcal{L}_{\mathrm{dis}}+\mathcal{L}_{\mathrm{gen}}$ & \ding{52} & \textbf{69.6} & \textbf{40.9} & \textbf{71.3} & \textbf{63.5} \\
\end{NiceTabular}
}%
};
\draw[line width=0.08pt, rounded corners=4pt]
  (tbl.south west) rectangle (tbl.north east);
\draw[line width=0.08pt, rounded corners=4pt, opacity=0.6]
  ([xshift=0.5pt,yshift=0.5pt]tbl.south west)
  rectangle
  ([xshift=0.5pt,yshift=0.5pt]tbl.north east);
\end{tikzpicture}
}
\vspace{-0.1in}
\end{table}

\subsection{Ablation on the Dual-Mode Embedder}
Table~\ref{tab:ablation_embedder} studies the two training signals of the embedder. Adding the generative objective $\mathcal{L}_{\mathrm{gen}}$ to the discriminative objective slightly improves the overall score from 62.3 to 62.6 without hard negatives. Although image performance remains nearly unchanged, video and visual-document performance increase by 0.7 and 0.9 points, respectively, suggesting that explicit supervision for RC-CoT-conditioned queries is especially useful for inputs requiring temporal or document-level reasoning. Introducing mined hard negatives further raises the overall score to 63.5, with gains of 1.0, 0.6, and 1.0 points on image, video, and visual-document tasks over dual-mode training without hard negatives. The complete embedder thus improves by 1.2 points overall over discriminative-only training. These results indicate that the two components play distinct roles: $\mathcal{L}_{\mathrm{gen}}$ teaches the embedder to convert RC-CoT into retrieval-effective representations, while hard negatives sharpen the decision boundary in both embedding modes.

\begin{table*}[t]
\centering
\caption{Ablation of retrieval-oriented reinforcement learning. We report the aggregated results on image, video, and visual-document tasks in MMEB-V2.}
\label{tab:ablation_rl}
\vspace{-0.05in}
\renewcommand{\arraystretch}{1}
\tabcolsep2pt
\resizebox{0.75\linewidth}{!}{
\begin{tikzpicture}
\node[inner sep=2.5pt] (tbl) {
{
\begin{NiceTabular}{lccc cccc}
\multirow{2}{*}{\textbf{Variant}} &
\multicolumn{3}{c}{\textbf{Retrieval Reward}} &
\multicolumn{4}{c}{\textbf{MMEB-V2}} \\
\cmidrule(lr){2-4} \cmidrule(lr){5-8}
& $\boldsymbol{R_{\mathrm{judge}}}$ &
$\boldsymbol{R_{\mathrm{rank}}}$ & $\boldsymbol{R_{\mathrm{cot}}}$ &
\textbf{Image} & \textbf{Video} & \textbf{VisDoc} & \textbf{Overall} \\
\midrule
SFT & \ding{56} & \ding{56} & \ding{56} & 72.1 & 52.7 & 76.4 & 68.8 \\
w/o judge reward & \ding{56} & \ding{52} & \ding{52} & 73.1 & 53.3 & \textbf{76.7} & 69.5 \\
w/o rerank reward & \ding{52} & \ding{56} & \ding{52} & 73.0 & 53.3 & 76.5 & 69.5 \\
w/o RC-CoT reward & \ding{52} & \ding{52} & \ding{56} & 73.3 & 53.0 & 76.5 & 69.6 \\
\textbf{Full GRPO} & \ding{52} & \ding{52} & \ding{52} & \textbf{73.4} & \textbf{53.9} & 76.6 & \textbf{69.9} \\
\end{NiceTabular}
}%
};
\draw[line width=0.08pt, rounded corners=4pt]
  (tbl.south west) rectangle (tbl.north east);
\draw[line width=0.08pt, rounded corners=4pt, opacity=0.6]
  ([xshift=0.5pt,yshift=0.5pt]tbl.south west)
  rectangle
  ([xshift=0.5pt,yshift=0.5pt]tbl.north east);
\end{tikzpicture}
}
\vspace{-0.1in}
\end{table*}

\subsection{Effect of Retrieval-Oriented Reinforcement Learning}
We evaluate retrieval-oriented reinforcement learning in Table~\ref{tab:ablation_rl}. Full GRPO improves the overall score from 68.8 for the SFT adviser to 69.9, a gain of 1.1 points, with consistent improvements on image and video tasks. Removing any outcome-oriented reward degrades end-to-end performance: excluding the judge or rerank reward lowers the overall score by 0.4 points, while excluding the RC-CoT reward causes a 0.3 points drop. The full model obtains the best image, video, and overall results. These results show that supervised imitation alone does not fully align adviser outputs with retrieval outcomes. The judge, rerank, and RC-CoT rewards provide complementary supervision for path selection, candidate ordering, and query correction, respectively, enabling GRPO to optimize the complete rerank-or-retrieve process. Please refer to Appendix~\ref{sec:detailed_routing_behavior} for more analysis.

\section{Analysis}

\subsection{Behavior Before and After GRPO}

\begin{figure}[t!]
    \centering
    \begin{minipage}[t]{0.55\linewidth}
        \centering
        
        \includegraphics[width=\linewidth]{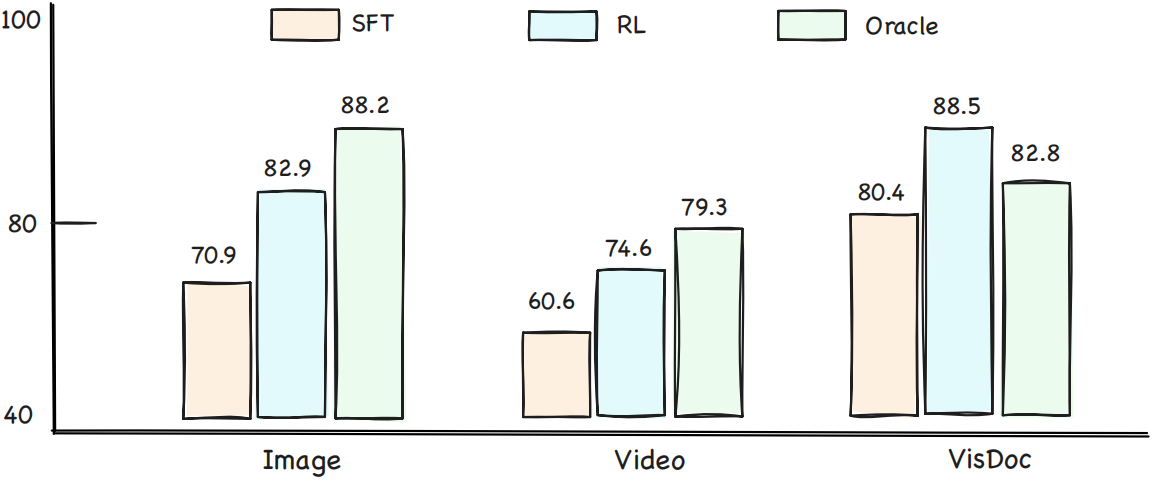}
        \vspace{-0.2in}
        \captionof{figure}{Routing calibration before and after GRPO. Bars show modality-level reranking rates, while GT@Top-5 indicates target availability in the initial candidates.}
        \vspace{-0.1in}
        \label{fig:grpo_behavior}
    \end{minipage}
    \hfill
    \hfill
    \begin{minipage}[t]{0.44\linewidth}
        \centering
        
        \includegraphics[width=\linewidth]{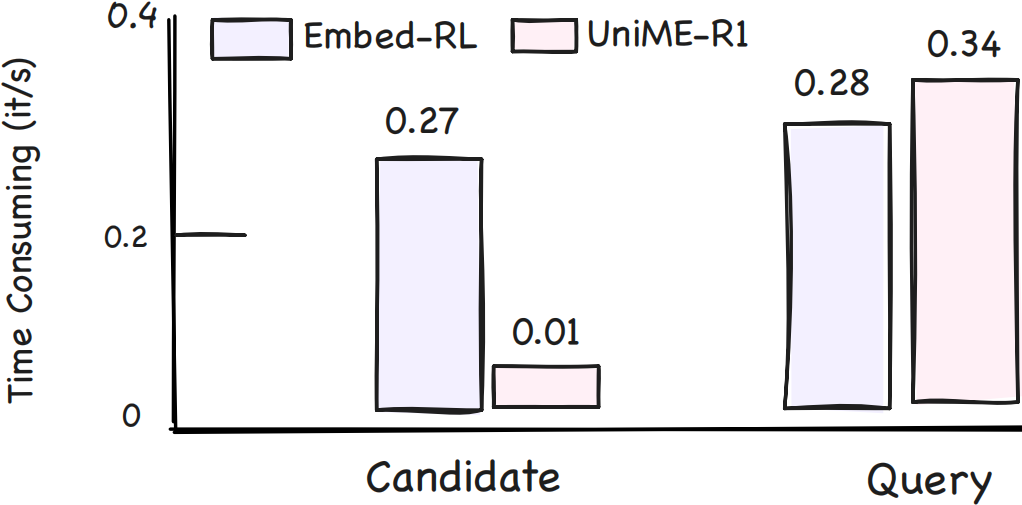}
        \vspace{-0.2in}
        \captionof{figure}{Average processing latency per query and candidate on MMEB-V1, measured over 3,600 queries and 111,384 candidates.}
        \vspace{-0.1in}
        \label{fig:latency}
    \end{minipage}
\end{figure}

We study how retrieval-oriented reinforcement learning calibrates the adviser’s routing policy. Figure~\ref{fig:grpo_behavior} reports the fraction of queries routed to reranking before and after GRPO, together with GT@Top-5, which measures target availability in the initial candidate set. GRPO reduces the routing–GT@Top-5 gap from 17.3 to 5.3 points on image tasks and from 18.7 to 4.7 points on video tasks. On VisDoc, the gap increases from 2.4 to 5.7 points, suggesting mild over-routing to reranking. These aggregate rates evaluate policy calibration rather than per-query judge accuracy. Detailed results are provided in the Appendix~\ref{sec:detailed_routing_behavior}.

\begin{table}[t]
\centering
\caption{Effect of multi-round inference on MMEB-V2.}
\label{tab:multi_round}
\vspace{-0.1in}
\renewcommand{\arraystretch}{1}
\tabcolsep3pt
\resizebox{0.65\linewidth}{!}{
\begin{tikzpicture}
\node[inner sep=2.5pt] (tbl) {
{
\begin{NiceTabular}{lc cccc}
\multirow{2}{*}{\textbf{Method}} &
\multirow{2}{*}{\textbf{Rounds}} &
\multicolumn{4}{c}{\textbf{MMEB-V2}} \\
\cmidrule(lr){3-6}
& & \textbf{Image} & \textbf{Video} & \textbf{VisDoc} & \textbf{Overall} \\
\midrule
UniME-R1 & 1 & 73.4 & 53.9 & 76.6 & 69.9 \\
\textbf{UniME-R1++} & 2 & \textbf{73.7} & \textbf{55.2} & \textbf{77.1} & \textbf{70.5} \\
\end{NiceTabular}
}%
};
\draw[line width=0.08pt, rounded corners=4pt]
  (tbl.south west) rectangle (tbl.north east);
\draw[line width=0.08pt, rounded corners=4pt, opacity=0.6]
  ([xshift=0.5pt,yshift=0.5pt]tbl.south west)
  rectangle
  ([xshift=0.5pt,yshift=0.5pt]tbl.north east);
\end{tikzpicture}
}
\vspace{-0.1in}
\end{table}

\subsection{Inference Efficiency}
We evaluate inference efficiency on MMEB-V1 with 3,600 queries and 111,384 candidates, reporting average latency per query or candidate in Figure~\ref{fig:latency}. Embed-RL generates CoT for both queries and candidates, costing 0.27 s per candidate. UniME-R1 generates RC-CoT only for queries and encodes each candidate once with \disemb, reducing candidate latency to 0.01 s ($27\times$ faster). This advantage is substantial for large or frequently updated candidate pools, where candidate-side cost dominates index construction and maintenance. On the query side, UniME-R1 increases latency modestly from 0.28 s to 0.34 s. Although retrieval-aware reasoning adds computation, the rerank-or-retrieve mechanism avoids unnecessary full-corpus retrieval when the target is already in the initial top-$k$, keeping average query overhead limited.

\subsection{Multi-Round Inference}
To assess the benefit of iterative retrieval feedback, we extend UniME-R1 from one to two inference rounds. As shown in Table~\ref{tab:multi_round}, after scaling the test time, UniME-R1++ improves the overall score from 69.9 to 70.5, including gains of 0.3, 1.3, and 0.5 points on image, video, and visual-document tasks, respectively. The largest gain appears on video retrieval, where an additional round better resolves fine-grained temporal ambiguity. These results indicate that UniME-R1 benefits from iterative refinement. Nevertheless, given the modest performance gain relative to the extra inference cost, we adopt a single round as the default setting.

\section{Conclusion}

In this paper, we introduce UniME-R1, an embedder--adviser framework that generates Retrieval-Centric Chain-of-Thought (RC-CoT) from initial retrieval feedback. By diagnosing the confusion exposed by retrieved candidates, UniME-R1 adaptively reranks the current candidates or performs RC-CoT-enhanced re-retrieval with a reusable index. Hard-negative training and retrieval-oriented reinforcement learning further align both components with retrieval outcomes. Experiment results demonstrate that UniME-R1 achieves consistent improvements on the MMEB-V2 benchmark and diverse general multimodal retrieval tasks, validating retrieval feedback as an effective basis for multimodal retrieval reasoning. We hope this work provides useful insights into universal multimodal representation learning.
\newpage
\appendix


\section{Detailed Dataset Construction}

\subsection{Hard Negative Mining}
\label{data_statistic}

Given a query--positive pair $(q,c^{+})$, we first use an off-the-shelf multimodal embedder (Qwen3-VL-Embedder 8B) to retrieve top-ranked candidates from $\mathcal{C}$. To reduce false negatives, we remove retrieved candidates whose similarity is overly close to or exceeds that of the positive candidate with a predefined threshold $\tau_h$: $s(q,c)-s(q,c^{+})>\tau_h$. We further score each remaining query--candidate pair using an LVLM judge constrained to answer ``\texttt{yes}'' or ``\texttt{no}'' using the following prompt:

\begin{promptblock}
\vspace{-2mm}
\noindent\textit{You are an expert data evaluator. Your task is to evaluate whether a 'Candidate' satisfies the requirements of a given 'Query'.\\
Decision Rules:\\
- Output 'Yes' if the Candidate satisfies the Query.\\
- Output 'No' if the Candidate does NOT satisfy the \\Query (e.g., missing attributes, wrong objects, completely irrelevant, etc.)\\
Output exactly one word: Yes or No.}
\vspace{-2mm}
\end{promptblock}

Instead of using the decoded answer, we define a soft relevance score from the answer logits:
$
    a(q,c)
    =
    \operatorname{logit}(\text{\texttt{yes}})
    -
    \operatorname{logit}(\text{\texttt{no}}).
$
We also compute $a(q,c^{+})$ for the target. Candidates are ranked by $a(q,c)$, and those satisfying $a(q,c)<a(q,c^{+})$ are retained as challenging but incorrect candidates. The resulting hard-negative pool $\mathcal{H}(q)$ is shared across contrastive learning, teacher annotation, and retrieval-oriented rewards.

\begin{table}[h!]
\centering
\caption{Statistics of the training data.}
\label{tab:adviser_data_stats}
\vspace{-0.1in}
\renewcommand{\arraystretch}{1}
\tabcolsep3pt
\resizebox{0.55\linewidth}{!}{
\begin{tikzpicture}
\node[inner sep=2.5pt] (tbl) {
{
\begin{NiceTabular}{lrrrr}
\textbf{Modality} & \textbf{SFT Pos.} & \textbf{SFT Neg.} & \textbf{SFT All} & \textbf{GRPO} \\
\midrule
Image  & 108,925 & 104,925 & 213,850 & 9,650 \\
VisDoc & 42,102  & 49,849  & 91,951  & 1,870 \\
Video  & 187,190 & 150,089 & 337,279 & 1,630 \\
\midrule
\textbf{Overall} & \textbf{338,217} & \textbf{304,863} & \textbf{643,080} & \textbf{13,150} \\
\end{NiceTabular}
}%
};
\draw[line width=0.08pt, rounded corners=4pt]
  (tbl.south west) rectangle (tbl.north east);
\draw[line width=0.08pt, rounded corners=4pt, opacity=0.6]
  ([xshift=0.5pt,yshift=0.5pt]tbl.south west)
  rectangle
  ([xshift=0.5pt,yshift=0.5pt]tbl.north east);
\end{tikzpicture}
}
\vspace{-0.1in}

\end{table}

\subsection{Data Sources and Modality-Balanced Sampling}

Following the training-data paradigm of VLM2Vec-V2~\citep{meng2025vlm2vecv2}, we construct a comprehensive multimodal retrieval corpus from three principal sources: (1) video--language instruction data from LLaVA-Hound~\citep{llava_hound}, (2) visual-document retrieval data from ViDoRe~\citep{vidore} and VisRAG~\citep{yu2025visrag}, and (3) image-based vision-task data from MMEB-train~\citep{jiang2024vlm2vec}. Together, these sources cover images, videos, and visual documents, as well as diverse retrieval intents such as classification, visual question answering, cross-modal retrieval, grounding, and temporal localization.

The source datasets vary substantially in scale. Directly combining all available examples would cause large video datasets to dominate training and weaken coverage of image and visual-document tasks. We therefore perform modality-aware stratified sampling and impose the following maximum size on each constituent dataset:
\begin{itemize}
    \item \textbf{Image datasets:} at most 50,000 training examples per dataset;
    \item \textbf{Visual-document datasets:} at most 100,000 training examples per dataset;
    \item \textbf{Video datasets:} at most 300,000 training examples per dataset.
\end{itemize}
These thresholds are applied before data filtering. Within each modality, examples are sampled from every constituent dataset to retain broad task coverage rather than allowing a small number of large datasets to determine the training distribution.

\subsection{Merging Image Classification Datasets}

MMEB-train contains several image-classification datasets with small label spaces. For example, HatefulMemes contains only 2 categories, VOC2007 contains 20, and N24News contains 24. When a contrastive-learning batch is sampled from one such dataset, multiple examples can share the same class semantics. Nevertheless, the standard in-batch contrastive objective treats candidates paired with other queries as negatives, producing many false negatives and conflicting supervision.

To reduce these collisions, we merge N24News, HatefulMemes, VOC2007, and SUN397 into a unified image-classification training set before batch construction. Sampling a batch from the merged set substantially increases label and semantic diversity, thereby reducing the probability that semantically equivalent image--text pairs are incorrectly contrasted as negatives. This operation changes only the batching pool; the original query--target correspondence and dataset-specific task instructions are preserved.

\begin{table}[t]
\centering
\caption{Effect of the number of retrieved candidates visible to the teacher adviser during zero-shot inference on MMEB-V2. We report Hit@1 for image and video tasks and nDCG@5 for visual-document tasks.}
\label{tab:ablation_context_size}
\vspace{-0.1in}
\renewcommand{\arraystretch}{1.05}
\tabcolsep4pt
\resizebox{0.5\linewidth}{!}{
\begin{tikzpicture}
\node[inner sep=2.5pt] (tbl) {
{
\begin{NiceTabular}{lcccc}
\multirow{2}{*}{\textbf{Top-k}} &
\multicolumn{4}{c}{\textbf{MMEB-V2}} \\
\cmidrule(lr){2-5}
& \textbf{Image}
& \textbf{Video}
& \textbf{VisDoc}
& \textbf{Overall} \\
\midrule
Top-1 & 71.7 & 45.7 & 72.4 & 65.9 \\
Top-3 & 73.3 & 52.4 & 75.3 & 69.1 \\
Top-5 & \textbf{73.4} & 53.9 & 76.6 & 69.9 \\
Top-7 & \textbf{73.4}  & \textbf{54.3} & \textbf{78.3} & \textbf{70.5} \\
\end{NiceTabular}
}%
};
\draw[line width=0.08pt, rounded corners=4pt]
  (tbl.south west) rectangle (tbl.north east);
\draw[line width=0.08pt, rounded corners=4pt, opacity=0.6]
  ([xshift=0.5pt,yshift=0.5pt]tbl.south west)
  rectangle
  ([xshift=0.5pt,yshift=0.5pt]tbl.north east);
\end{tikzpicture}
}
\vspace{-0.1in}
\end{table}

\subsection{Data Assignment for the Dual-Mode Embedder}

The two query representations of the embedder have different supervision requirements. The discriminative representation \disemb\ must retain broad multimodal retrieval ability and is therefore trained on all examples retained after modality-balanced sampling. In contrast, training \genemb\ requires a reliable teacher-generated RC-CoT. We consequently apply the generative contrastive objective only to examples that pass the RC-CoT quality filters.

Let $\mathcal{D}_{\mathrm{all}}$ denote the complete sampled corpus and let $\mathcal{D}_{\mathrm{cot}}\subseteq\mathcal{D}_{\mathrm{all}}$ denote the subset with valid RC-CoT. The per-example embedder objective can be written as
\begin{equation}
    \mathcal{L}_{\mathrm{emb}}(q_i)
    =
    \mathcal{L}_{\mathrm{dis}}(q_i)
    +
    \mathbb{1}[q_i\in\mathcal{D}_{\mathrm{cot}}]
    \mathcal{L}_{\mathrm{gen}}(q_i).
\end{equation}
This selective assignment prevents low-quality reasoning traces from degrading \genemb\ while preserving the full data coverage of \disemb\.
Table~\ref{tab:adviser_data_stats} summarizes the data used for the SFT stages. We uses 643,080 structured examples, comprising 338,217 positive and 304,863 negative instances.

\subsection{RL Data Sampling}

We construct the reinforcement-learning training dataset from the remaining difficult examples. Because the challenging sub-datasets differ considerably in size, naive random sampling would bias training toward the largest sources. We therefore apply equidistant sampling within each sub-dataset and control its contribution to the final RL set. This exposes GRPO to diverse retrieval failures rather than repeatedly optimizing a few dominant tasks. The resulting modality-level sample counts are reported in Table~\ref{tab:adviser_data_stats}.

\section{Additional Ablation Studies}
\subsection{Effect of Retrieved Context Size}

We study how the number of retrieved candidates visible to the adviser affects retrieval-oriented reasoning. Following the RC-CoT ablation, we use the teacher model for zero-shot inference and keep all other settings unchanged. The only variable is the number of top-ranked candidates provided as retrieval context.

As shown in Table~\ref{tab:ablation_context_size}, expanding the visible context from Top-1 to Top-3 improves the overall score by 3.2 points, with the largest gain of 6.7 points on video tasks. Increasing the context to Top-5 and Top-7 yields further improvements, reaching the best overall score of 70.5. Image performance saturates at Top-5, whereas video and VisDoc tasks continue to benefit from additional candidates. These results suggest that broader retrieval context provides useful comparative evidence for resolving temporal and document-level ambiguity, although the diminishing gains beyond Top-5 indicate a trade-off between effectiveness and inference cost.

\subsection{Effect of Candidate-Wise Reranking Analysis}
The \rerankthink\ field requires the adviser to analyze the relevance of each retrieved candidate before predicting the candidate order and routing decision. To examine whether this intermediate reasoning is necessary, we compare the full adviser with a variant that omits \rerankthink\ while retaining the remaining outputs.

\begin{table}[t]
\centering
\caption{Effect of candidate-wise analysis in \texttt{<rerank\_think>} on MMEB-V2.}
\label{tab:ablation_think}
\vspace{-0.05in}
\renewcommand{\arraystretch}{1}
\tabcolsep2pt
\resizebox{0.7\linewidth}{!}{
\begin{tikzpicture}
\node[inner sep=2.5pt] (tbl) {
{
\begin{NiceTabular}{lc cccc}
\multirow{2}{*}{\textbf{Variant}} &
\multirow{2}{*}{\rerankthink} &
\multicolumn{4}{c}{\textbf{MMEB-V2}} \\
\cmidrule(lr){3-6}
& & \textbf{Image} & \textbf{Video} & \textbf{VisDoc} & \textbf{Overall} \\
\midrule
w/o think & \ding{55} & 71.6 & 47.5 & 75.7 & 67.3 \\
Full & \ding{51} & \textbf{73.4} & \textbf{53.9} & \textbf{76.6} & \textbf{69.9} \\
\end{NiceTabular}
}%
};
\draw[line width=0.08pt, rounded corners=4pt]
  (tbl.south west) rectangle (tbl.north east);
\draw[line width=0.08pt, rounded corners=4pt, opacity=0.6]
  ([xshift=0.5pt,yshift=0.5pt]tbl.south west)
  rectangle
  ([xshift=0.5pt,yshift=0.5pt]tbl.north east);
\end{tikzpicture}
}
\vspace{-0.1in}
\end{table}

As shown in Table~\ref{tab:ablation_think}, removing candidate-wise analysis reduces the overall score from 69.9 to 67.3. The full model improves image and visual-document performance by 1.8 and 0.9 points, respectively, while the gain reaches 6.4 points on video tasks. This larger improvement suggests that explicit candidate comparison is particularly useful for resolving fine-grained temporal and event-level differences. Overall, \rerankthink\ serves as a useful intermediate reasoning step rather than merely an explanatory output, supporting more reliable candidate ordering and routing.

\section{Detailed Routing Behavior Before and After GRPO}
\label{sec:detailed_routing_behavior}

Table~\ref{tab:grpo_behavior_detailed} provides the task-level routing statistics underlying Figure~\ref{fig:grpo_behavior}. GRPO raises the reranking rate for every task group, but this increase should be interpreted as a policy shift rather than an improvement by itself. Relative to GT@Top-5, the GRPO routing frequency becomes closer in nine of the twelve task groups; the exceptions are Video-CLS, Video-RET, and VisDoc-VDRv1. At the modality level, GRPO substantially reduces the aggregate gap on image and video tasks, whereas it moves the VisDoc reranking rate beyond GT@Top-5. These frequency-level comparisons measure aggregate calibration and do not establish whether the correct individual queries are routed.

\begin{table}[t]
\centering
\caption{Detailed routing behavior before and after GRPO. GT@Top-5, SFT, GRPO, and $\Delta$ are percentages; $\Delta$ denotes GRPO minus SFT. Average rows are weighted by the number of datasets.}
\vspace{-0.05in}
\label{tab:grpo_behavior_detailed}
\renewcommand{\arraystretch}{1}
\tabcolsep2pt
\resizebox{0.7\linewidth}{!}{
\begin{tikzpicture}
\node[inner sep=2.5pt] (tbl) {
{
\begin{NiceTabular}{llccccc}
\textbf{Modality} & \textbf{Task Group} & \textbf{\# Datasets} & \textbf{GT@Top-5} & \textbf{SFT} & \textbf{GRPO} & \textbf{$\Delta$} \\
\midrule
\multirow{5}{*}{Image}
 & CLS  & 10 & 86.2 & 77.8 & 89.1 & +11.3 \\
 & QA   & 10 & 85.7 & 81.9 & 86.9 & +5.0 \\
 & RET  & 12 & 89.4 & 54.9 & 72.9 & +18.0 \\
 & GD   & 4  & 95.7 & 74.0 & 87.4 & +13.4 \\
 & \textbf{Avg.} & \textbf{36} & \textbf{88.2} & \textbf{70.9} & \textbf{82.9} & \textbf{+12.0} \\
\midrule
\multirow{5}{*}{Video}
 & CLS  & 5 & 70.3 & 64.3 & 78.2 & +13.9 \\
 & QA   & 5 & 100.0 & 78.0 & 85.3 & +7.3 \\
 & RET  & 5 & 48.9 & 44.3 & 64.6 & +20.3 \\
 & MRET & 3 & 81.4 & 52.5 & 67.3 & +14.8 \\
 & \textbf{Avg.} & \textbf{18} & \textbf{79.3} & \textbf{60.6} & \textbf{74.6} & \textbf{+14.0} \\
\midrule
\multirow{5}{*}{VisDoc}
 & VDRv1 & 10 & 75.9 & 88.8 & 94.1 & +5.3 \\
 & VDRv2 & 4  & 76.5 & 65.0 & 80.0 & +15.0 \\
 & VR    & 6  & 90.1 & 80.2 & 89.0 & +8.8 \\
 & OOD   & 4  & 84.7 & 74.8 & 82.4 & +7.6 \\
 & \textbf{Avg.} & \textbf{24.0} & \textbf{82.8} & \textbf{80.4} & \textbf{88.5} & \textbf{+8.1} \\
\end{NiceTabular}
}%
};
\draw[line width=0.08pt, rounded corners=4pt]
  (tbl.south west) rectangle (tbl.north east);
\draw[line width=0.08pt, rounded corners=4pt, opacity=0.6]
  ([xshift=0.5pt,yshift=0.5pt]tbl.south west)
  rectangle
  ([xshift=0.5pt,yshift=0.5pt]tbl.north east);
\end{tikzpicture}
}
\vspace{-0.1in}
\end{table}

\section{Gains After RL}
To isolate the contribution of retrieval-oriented reinforcement learning, Table~\ref{tab:rl_gain_detailed} compares identical inference configurations using the SFT and RL advisers. The baseline remains unchanged because it does not use adviser outputs. When reranking and RC-CoT are both enabled, RL improves the overall score from 68.8 to 69.9 ($+1.1$), including gains of 1.3 points on image tasks and 0.9 points on video tasks.

\begin{table*}[t]
\centering
\caption{Detailed comparison of SFT and RL advisers under different inference configurations on MMEB-V2. Results are grouped into image, video, and visual-document (VisDoc) tasks, and $\Delta$ reports the RL minus SFT change in the overall score. \emph{RC-CoT only} and \emph{Rerank only} force a single inference path.}
\vspace{-0.05in}
\label{tab:rl_gain_detailed}
\renewcommand{\arraystretch}{1.08}
\tabcolsep3.5pt
\resizebox{\textwidth}{!}{
\begin{tikzpicture}
\node[inner sep=2.5pt] (tbl) {
{
\begin{NiceTabular}{l cccc cccc c}
\multirow{2}{*}{\textbf{Inference Configuration}} &
\multicolumn{4}{c}{\textbf{SFT Adviser}} &
\multicolumn{4}{c}{\textbf{RL Adviser}} &
\multirow{2}{*}{\textbf{$\Delta$ Overall}} \\
\cmidrule(lr){2-5} \cmidrule(lr){6-9}
& \textbf{Image} & \textbf{Video} & \textbf{VisDoc} & \textbf{Overall}
& \textbf{Image} & \textbf{Video} & \textbf{VisDoc} & \textbf{Overall} & \\
\midrule
Baseline & 69.7 & 40.7 & 71.6 & 63.6 & 69.7 & 40.7 & 71.6 & 63.6 & +0.0 \\
+ Rerank & 71.8 & 50.9 & 75.9 & 68.3 & 72.8 & 51.9 & 75.8 & 68.9 & +0.6 \\
+ Rerank + RC-CoT & \textbf{72.1} & 53.0 & \textbf{76.8} & \textbf{68.8} & \textbf{73.4} & \textbf{53.9} & \textbf{76.6} & \textbf{69.9} & +1.1 \\
RC-CoT only & 71.0 & 50.5 & 75.9 & 67.8 & 72.8 & 52.2 & 76.0 & 69.0 & \textbf{+1.2} \\
Rerank only & 71.5 & \textbf{53.7} & 76.3 & 68.5 & 72.6 & 53.4 & 76.1 & 69.2 & +0.7 \\
\end{NiceTabular}
}%
};
\draw[line width=0.08pt, rounded corners=4pt]
  (tbl.south west) rectangle (tbl.north east);
\draw[line width=0.08pt, rounded corners=4pt, opacity=0.6]
  ([xshift=0.5pt,yshift=0.5pt]tbl.south west)
  rectangle
  ([xshift=0.5pt,yshift=0.5pt]tbl.north east);
\end{tikzpicture}
}
\vspace{-0.2in}
\end{table*}

The largest improvement occurs in the RC-CoT-only setting, which gains 1.2 points overall, including 1.8 points on image tasks and 1.7 points on video tasks. By contrast, the rerank-only setting improves by 0.7 points overall and slightly decreases on video and VisDoc. This asymmetry indicates that RL primarily strengthens retrieval-failure diagnosis and query refinement rather than uniformly improving candidate ordering. Moreover, the full adaptive model reaches 69.9, outperforming both forced RC-CoT-only (69.0) and rerank-only (69.2) inference. The gain therefore comes not only from stronger individual actions, but also from better coordination between reranking and re-retrieval.

\section{Comprehensive Performance on Video Retrieval}
\label{app:uvrb}

\subsection{UVRB Evaluation Metrics}
\label{uvab:metrics}
We evaluate video retrieval generalization on the Universal Video Retrieval Benchmark (UVRB)~\citep{guo2025towards}. UVRB assesses video retrieval models along three orthogonal dimensions—Tasks, Domains, and Sub-domains—and uses unweighted arithmetic means to ensure fair comparison across heterogeneous datasets. It comprises 16 datasets, exhaustively partitioned into non-overlapping categories along these dimensions (Table~\ref{tab:uvrb_partition}), providing a structured basis for analyzing model performance in diverse retrieval scenarios. Following the UVRB protocol, we report R@1 by default, R@10 for CMRB and LoVR-TH, and P@1 for the multi-positive MS-TI and MS-TV datasets. The overall AVG score in Table~\ref{tab:zero-shot-retrieval} is computed as the arithmetic mean over the three task groups (TXT, CMP, VIS) and the three domain groups (CG, FG, LC), thereby summarizing performance across core retrieval paradigms rather than averaging directly over datasets.

\begin{table*}[t]
\centering
\caption{Detailed Partition of Datasets in the Universal Video Retrieval Benchmark (UVRB) Across Tasks, Domains, and Sub-domains.}
\label{tab:uvrb_partition}
\vspace{-0.05in}
\renewcommand{\arraystretch}{1}
\tabcolsep3pt
\resizebox{0.9\linewidth}{!}{
\begin{tikzpicture}
\node[inner sep=2.5pt] (tbl) {
{
\begin{NiceTabular}{l>{\raggedright\arraybackslash}p{0.78\linewidth}}
\textbf{Partition} & \textbf{Content} \\
\midrule
$\mathcal{D}_{\mathrm{TXT}}$ 
& \{MSRVTT, DiDeMo, CRB-G, CRB-S, VDC-O, CRB-T, CMRB, DREAM-E, LoVR-TH, PEV-K, LoVR-V, VDC-D\} \\

$\mathcal{D}_{\mathrm{CMP}}$ 
& \{MS-TI, MS-TV\} \\

$\mathcal{D}_{\mathrm{VIS}}$ 
& \{MSRVTT-I2V, LoVR-C2V\} \\

$\mathcal{D}_{\mathrm{CG}}$ 
& \{MSRVTT, DiDeMo, CRB-G\} \\

$\mathcal{D}_{\mathrm{FG}}$ 
& \{CRB-S, VDC-O, CRB-T, CMRB, DREAM-E, LoVR-TH, PEV-K\} \\

$\mathcal{D}_{\mathrm{LC}}$ 
& \{LoVR-V, VDC-D\} \\

$\mathcal{D}_{\mathrm{S}}$ 
& \{CRB-S, VDC-O\} \\

$\mathcal{D}_{\mathrm{T}}$ 
& \{CRB-T, CMRB\} \\

$\mathcal{D}_{\mathrm{PR}}$ 
& \{DREAM-E, LoVR-TH, PEV-K\} \\
\end{NiceTabular}
}%
};
\draw[line width=0.08pt, rounded corners=4pt]
  (tbl.south west) rectangle (tbl.north east);
\draw[line width=0.08pt, rounded corners=4pt, opacity=0.6]
  ([xshift=0.5pt,yshift=0.5pt]tbl.south west)
  rectangle
  ([xshift=0.5pt,yshift=0.5pt]tbl.north east);
\end{tikzpicture}
}
\vspace{-0.1in}
\end{table*}

\subsection{Per-Dataset Results}
Table~\ref{tab:uvrb_datasets} presents the full per-dataset comparison. UniME-R1 achieves average scores of 58.2 and 58.7 with the 2B and 4B embedders, respectively, outperforming the corresponding Embed-RL models by 1.0 and 1.3 points under the same 16-dataset evaluation protocol. The gains are particularly pronounced on composed retrieval. For example, the 2B variant achieves 50.5 on MS-TI and 43.8 on MS-TV, substantially surpassing Embed-RL-2B, which obtains 19.3 and 21.0 on the same datasets. UniME-R1 also performs strongly on temporal retrieval and challenging video-to-video matching, suggesting that the benefits of retrieval feedback generalize beyond the MMEB-V2 setting.

\begin{table*}[t]
\centering
\caption{Per-dataset video retrieval results on UVRB. AVG is the arithmetic mean over all 16 datasets. Metrics are R@1 unless otherwise indicated; the best and second-best results are bolded and underlined.}
\vspace{-0.05in}
\label{tab:uvrb_datasets}
\renewcommand{\arraystretch}{1.05}
\tabcolsep2.5pt
\resizebox{\textwidth}{!}{
\begin{tikzpicture}
\node[inner sep=2.5pt] (tblA) {
{
\begin{NiceTabular}{l c cccccccc}
\textbf{Model} & \textbf{AVG} & \textbf{MSRVTT} & \textbf{DiDeMo} & \textbf{CRB-G} & \textbf{CRB-S} & \textbf{VDC-O} & \textbf{CRB-T} & \textbf{CMRB} & \textbf{DREAM-E} \\
& & R@1 & R@1 & R@1 & R@1 & R@1 & R@1 & R@10 & R@1 \\
\midrule
CLIP4Clip~\citep{luo2022clip4clip} & 39.0 & 33.3 & 29.7 & 51.1 & 49.7 & 62.0 & 28.9 & 28.0 & 19.1 \\
ViCLIP~\citep{wang2023internvid} & 35.2 & 38.6 & 30.6 & 44.7 & 43.7 & 53.0 & 34.9 & 22.9 & 23.5 \\
VideoCLIP-XL~\citep{wang2024videoclip} & 49.1 & 44.3 & 40.3 & 82.8 & 83.9 & 73.5 & 48.7 & 27.4 & 26.3 \\
LanguageBind~\citep{zhu2024languagebind} & 48.7 & \underline{47.9} & 42.1 & 71.6 & 68.7 & 75.9 & 46.6 & 29.0 & 28.0 \\
InternVideo2-1B~\citep{wang2024internvideo2} & 40.4 & 44.9 & 40.4 & 58.6 & 56.8 & 64.4 & 47.0 & 35.5 & 24.2 \\
InternVideo2-6B~\citep{wang2024internvideo2} & 42.7 & \textbf{48.5} & 41.8 & 60.8 & 61.2 & 65.0 & 45.5 & 34.6 & 27.1 \\
GME-2B~\citep{zhang2025bridging} & 48.8 & 39.0 & 30.3 & 69.0 & 71.8 & 71.5 & 40.0 & 29.8 & 24.0 \\
Unite-2B~\citep{kong2025modality} & 48.0 & 36.7 & 29.8 & 69.9 & 72.3 & 72.7 & 40.9 & 28.4 & 22.3 \\
VLM2Vec-V2~\citep{meng2025vlm2vecv2} & 50.8 & 33.0 & 29.9 & 82.8 & 84.3 & 77.5 & 41.0 & 28.6 & 22.8 \\
BGE-VL~\citep{zhou2024megapairs} & 44.3 & 33.7 & 31.8 & 69.0 & 68.8 & 63.9 & 35.9 & 22.5 & 21.2 \\
UniME-7B~\citep{gu2025unime} & 52.1 & 35.1 & 33.5 & 81.5 & 82.7 & 74.3 & 47.6 & 31.7 & 29.3 \\
B3-7B~\citep{thirukovalluru2026breaking} & 51.1 & 28.2 & 35.0 & 81.5 & 82.5 & 76.8 & 41.5 & 31.2 & 21.6 \\
GME-7B~\citep{zhang2025bridging} & 53.0 & 43.6 & 37.7 & 74.0 & 76.7 & 73.1 & 44.2 & 30.4 & 27.4 \\
Unite-7B~\citep{kong2025modality} & 53.8 & 43.9 & 38.6 & 79.8 & 80.4 & 75.3 & 47.2 & 35.1 & 27.9 \\
GVE-3B~\citep{guo2025towards} & 54.4 & 43.1 & 37.6 & 85.0 & 84.6 & 78.6 & 49.6 & 36.3 & 28.0 \\
GVE-7B~\citep{guo2025towards} & 57.3 & 46.4 & \underline{43.3} & 86.5 & 84.7 & 79.4 & 53.9 & \textbf{39.8} & 30.2 \\
Embed-RL-2B~\citep{jiang2026embed} & 57.1 & 43.7 & 42.2 & \underline{91.4} & 89.7 & 84.6 & 49.4 & 36.5 & 31.8 \\
Embed-RL-4B~\citep{jiang2026embed} & 57.4 & 44.0 & \textbf{46.1} & \textbf{92.1} & \underline{89.9} & \textbf{85.9} & 53.8 & \underline{38.2} & 32.0 \\
\midrule
\textbf{UniME-R1-2B} & \underline{58.2} & 43.6 & 42.9 & 91.1 & \underline{89.9} & 83.4 & \underline{55.5} & 34.3 & \underline{33.3} \\
\textbf{UniME-R1-4B} & \textbf{58.7} & 45.3 & 43.2 & \textbf{92.1} & \textbf{90.4} & \underline{85.0} & \textbf{56.2} & 35.6 & \textbf{33.7} \\
\end{NiceTabular}
}%
};
\draw[line width=0.08pt, rounded corners=4pt] (tblA.south west) rectangle (tblA.north east);
\draw[line width=0.08pt, rounded corners=4pt, opacity=0.6]
  ([xshift=0.5pt,yshift=0.5pt]tblA.south west) rectangle ([xshift=0.5pt,yshift=0.5pt]tblA.north east);
\end{tikzpicture}
}
\vspace{1pt}

\resizebox{\textwidth}{!}{
\begin{tikzpicture}
\node[inner sep=2.5pt] (tblB) {
{
\begin{NiceTabular}{l cccccccc}
\textbf{Model} & \textbf{LoVR-TH} & \textbf{PEV-K} & \textbf{LoVR-V} & \textbf{VDC-D} & \textbf{MS-TI} & \textbf{MS-TV} & \textbf{MSRVTT-I2V} & \textbf{LoVR-C2V} \\
& R@10 & R@1 & R@1 & R@1 & P@1 & P@1 & R@1 & R@1 \\
\midrule
CLIP4Clip~\citep{luo2022clip4clip} & 33.8 & 17.9 & 36.0 & 56.6 & 17.3 & 18.3 & \textbf{92.4} & 50.3 \\
ViCLIP~\citep{wang2023internvid} & 20.2 & 7.5 & 23.0 & 39.5 & 28.3 & 24.3 & 84.6 & 43.3 \\
VideoCLIP-XL~\citep{wang2024videoclip} & 43.9 & 22.9 & 38.0 & 82.0 & 23.0 & 22.3 & 86.1 & 40.3 \\
LanguageBind~\citep{zhu2024languagebind} & 42.5 & 30.3 & 54.0 & 67.9 & 22.8 & 23.3 & 82.7 & 46.3 \\
InternVideo2-1B~\citep{wang2024internvideo2} & 29.8 & 2.6 & 28.0 & 48.5 & 26.5 & 23.0 & 79.4 & 36.8 \\
InternVideo2-6B~\citep{wang2024internvideo2} & 30.2 & 8.6 & 33.0 & 51.6 & 23.5 & 20.5 & 86.8 & 45.2 \\
GME-2B~\citep{zhang2025bridging} & 44.6 & 35.4 & 53.0 & 83.9 & 35.0 & 34.0 & 82.7 & 36.6 \\
Unite-2B~\citep{kong2025modality} & 44.5 & 35.5 & 57.0 & 79.2 & 25.0 & 23.3 & 86.3 & 44.5 \\
VLM2Vec-V2~\citep{meng2025vlm2vecv2} & 49.2 & 32.4 & 61.0 & 91.3 & 27.5 & 25.0 & 84.1 & 38.5 \\
BGE-VL~\citep{zhou2024megapairs} & 38.7 & 18.4 & 55.0 & 72.2 & 30.3 & 23.3 & 77.9 & 46.5 \\
UniME-7B~\citep{gu2025unime} & 50.4 & 32.3 & 48.0 & 84.7 & 31.0 & 30.5 & 86.7 & \textbf{53.7} \\
B3-7B~\citep{thirukovalluru2025breaking} & 46.2 & 38.7 & 59.0 & 85.3 & 27.5 & 26.5 & 88.4 & 47.1 \\
GME-7B~\citep{zhang2025bridging} & 52.3 & 39.6 & 71.0 & 86.5 & 34.8 & 33.3 & 86.0 & 37.0 \\
Unite-7B~\citep{kong2025modality} & 55.5 & \textbf{44.0} & 62.0 & 87.1 & 27.8 & 23.0 & 88.3 & 44.8 \\
GVE-3B~\citep{guo2025towards} & 52.2 & 33.0 & 61.0 & 91.8 & 34.0 & 26.8 & 89.1 & 40.3 \\
GVE-7B~\citep{guo2025towards} & 54.2 & \underline{41.3} & 68.0 & \underline{94.8} & 34.3 & 28.0 & 89.9 & 41.5 \\
Embed-RL-2B~\citep{jiang2026embed} & \underline{56.5} & 33.6 & \textbf{80.0} & 93.8 & 19.3 & 21.0 & 89.1 & \underline{51.4} \\
Embed-RL-4B~\citep{jiang2026embed} & \textbf{57.9} & 31.9 & \underline{77.0} & \textbf{95.2} & 15.8 & 21.0 & 87.9 & 49.0 \\
\midrule
\textbf{UniME-R1-2B} & 53.4 & 29.2 & 56.0 & 93.8 & \textbf{50.5} & \textbf{43.8} & 89.3 & 40.7 \\
\textbf{UniME-R1-4B} & 54.9 & 28.8 & 57.0 & 94.7 & \underline{48.8} & \underline{42.5} & \underline{91.0} & 40.0 \\
\end{NiceTabular}
}%
};
\draw[line width=0.08pt, rounded corners=4pt] (tblB.south west) rectangle (tblB.north east);
\draw[line width=0.08pt, rounded corners=4pt, opacity=0.6]
  ([xshift=0.5pt,yshift=0.5pt]tblB.south west) rectangle ([xshift=0.5pt,yshift=0.5pt]tblB.north east);
\end{tikzpicture}
}
\vspace{-0.2in}
\end{table*}

\begin{figure*}[t!]
    \centering
    
    \includegraphics[width=0.75\linewidth]{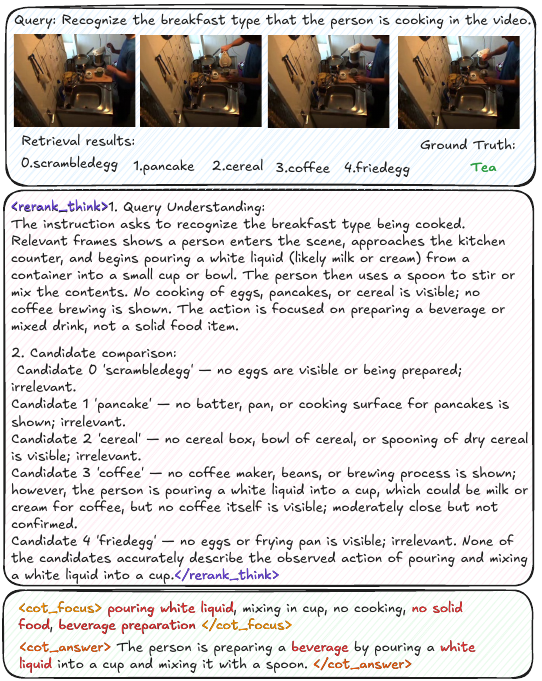}
    \caption{Case study of adviser-generated RC-CoT. Candidate-level analysis identifies corrective cues (\textit{e.g.}, the target contains no solid object) and guides re-retrieval toward the relevant white object.}
    \label{fig:case1}
    
    \vspace{-0.2in}
\end{figure*}

\begin{figure*}[t!]
    \centering
    \includegraphics[width=0.75\linewidth]{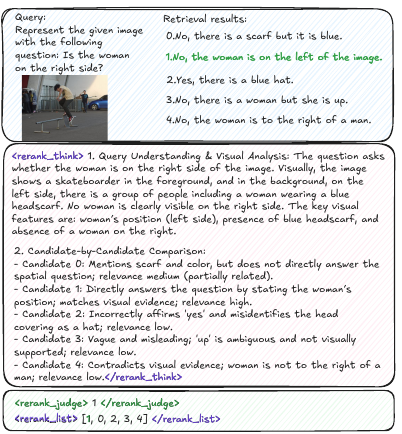}
    \vspace{-0.05in}
    \caption{Case study of adviser-guided reranking. Candidate-level relevance analysis separates the ground-truth target from a strong distractor, yielding a more accurate ranking.}
    \label{fig:case2}
    \vspace{-0.1in}
\end{figure*}

\section{Qualitative Case Studies}
Figures~\ref{fig:case1} and~\ref{fig:case2} illustrate how candidate-wise analysis supports the two inference paths. In Figure~\ref{fig:case1}, comparisons against retrieved candidates expose corrective evidence absent from the original query: the target contains no solid object and should instead emphasize the white object. Encoding these cues in RC-CoT provides a more targeted direction for re-retrieval. In Figure~\ref{fig:case2}, Candidate 4 is a strong distractor because it partially matches the queried relation and mentions a man who is indeed visible. By grounding each candidate in both the query intent and visual evidence, the adviser separates this distractor from the ground truth and produces the correct order.

\begin{figure*}[t!]
    \centering
    
    \includegraphics[width=\linewidth]{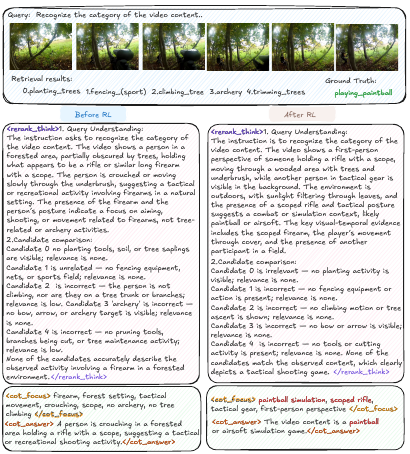}
    \vspace{-0.05in}
    \caption{Comparison of adviser-generated RC-CoT before and after RL. RL yields a more concise and accurate refinement focused on retrieval-critical evidence.}
    \label{fig:case3}
    \vspace{-0.2in}
\end{figure*}

\begin{figure*}[t!]
    \centering
    
    \includegraphics[width=\linewidth]{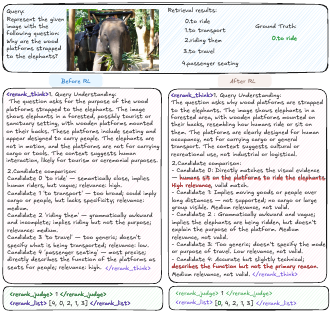}
    \vspace{-0.1in}
    \caption{Comparison of adviser-generated reranking before and after RL. RL improves alignment between candidate-level relevance judgments and query intent, yielding the correct ranking.}
    \label{fig:case4}
    \vspace{-0.2in}
    
\end{figure*}

\section{Comparative Examples of Adviser Before and After RL}
Figures~\ref{fig:case3} and~\ref{fig:case4} compare the SFT and RL advisers on the same inputs. In Figure~\ref{fig:case3}, the SFT-generated RC-CoT is broad and includes incidental context. After RL, the reasoning becomes shorter, more precise, and centered on the evidence needed to redirect retrieval. Figure~\ref{fig:case4} further shows improved intent alignment during reranking. Specifically, the RL adviser recognizes that \emph{passenger seating} describes a general function, whereas \emph{to ride} directly answers the purpose-oriented query. This distinction moves the ground-truth candidate above a semantically related distractor, showing that RL improves not merely response style but retrieval-oriented decision quality.

\section{Prompt for Inference}
\begin{figure*}[t!]
    \centering
    
    \includegraphics[width=\linewidth]{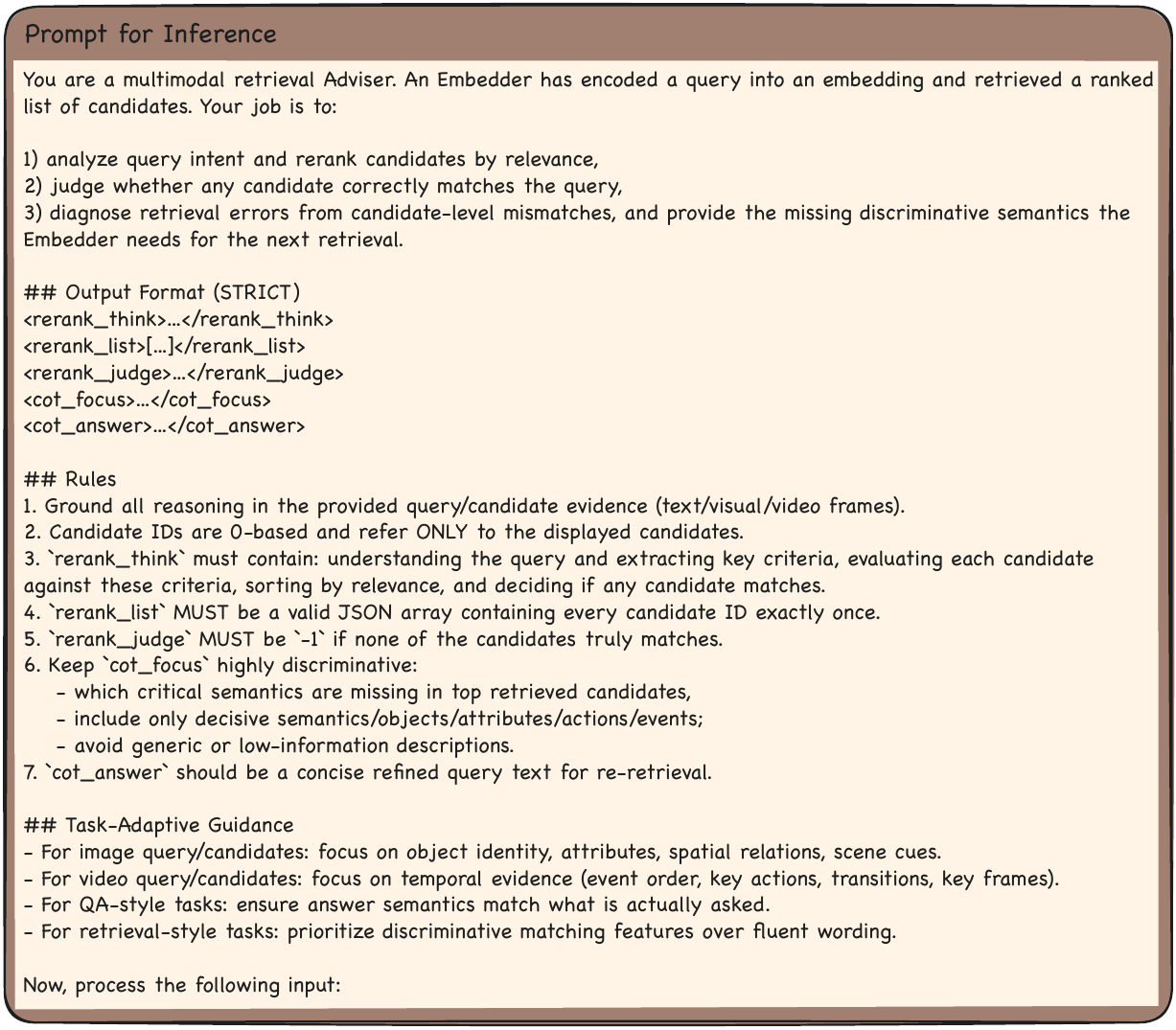}
    \vspace{-0.1in}
    \caption{Comparison of adviser-generated reranking before and after RL. RL improves alignment between candidate-level relevance judgments and query intent, yielding the correct ranking.}
    \label{fig:prompt_inference}
    \vspace{-0.2in}
    
\end{figure*}

Figure~\ref{fig:prompt_inference} shows the prompt we used in the inference process for Adviser.

\clearpage
{
\bibliography{UniME-R1}
\bibliographystyle{colm2024_conference}
}

\end{document}